\documentclass[conference]{IEEEtran}

\usepackage[export]{adjustbox}[2011/08/13]

\usepackage[flushleft]{threeparttable} 
\usepackage{booktabs}

\usepackage{graphicx}  
\usepackage[caption=false,font=footnotesize]{subfig}  
 
\usepackage{amsmath,amssymb,amsfonts}   
\usepackage{mathtools}
\usepackage{algorithm}
\usepackage{algorithmic}
\usepackage{textcomp}
\usepackage{xcolor}
\usepackage{array} 
\usepackage{multirow} 
\usepackage{enumerate} 

\usepackage{tabularx}
\usepackage{url}

\usepackage[export]{adjustbox}[2011/08/13]

\usepackage[flushleft]{threeparttable} 
\usepackage{booktabs}

\usepackage{graphicx}  
\usepackage[caption=false,font=footnotesize]{subfig}  
 
\usepackage{amsmath,amssymb,amsfonts}   
\usepackage{mathtools}
\usepackage{algorithm}
\usepackage{algorithmic}
\usepackage{textcomp}
\usepackage{xcolor}
\usepackage{array} 
\usepackage{multirow} 
\usepackage{enumerate} 

\usepackage{tabularx}
\usepackage{upgreek} 
\usepackage{stmaryrd} 

\makeatletter
\newcommand*\titleheader[1]{\gdef\@titleheader{#1}}
\AtBeginDocument{%
  \let\st@red@title\@title
  \def\@title{%
    \bgroup\normalfont\large\centering\@titleheader\par\egroup
    \vskip1.5em\st@red@title}
}
\makeatother

\title{Intelligently Augmented Contrastive Tensor Factorization: Empowering Multi-dimensional Time Series Classification in Low-Data Environments}
\titleheader{This manuscript has been accepted in Expert Systems with Applications. \\ DOI: \textbf{https://doi.org/10.1016/j.eswa.2025.127889} \\ \copyright~2025. This manuscript version is made available under the CC-BY-NC-ND 4.0 license \url{https://creativecommons.org/licenses/by-nc-nd/4.0/}
}

\author{
  \IEEEauthorblockN{Anushiya Arunan}
  \IEEEauthorblockA{
    Engineering Product Development\\
    Singapore University of Technology and Design\\
    Email: anushiya\_arunan@mymail.sutd.edu.sg
  }
  \and
  \IEEEauthorblockN{Yan Qin}
  \IEEEauthorblockA{
    School of Automation\\
    Chongqing University\\
    Email: yan.qin@cqu.edu.cn
  }
  \hspace{4.5cm} 
  \and
  \IEEEauthorblockN{Xiaoli Li}
  \IEEEauthorblockA{
    Institute for Infocomm Research\\
    Agency for Science, Technology and Research\\
    Email: xlli@i2r.a-star.edu.sg
  }
  
    \and
  \IEEEauthorblockN{Chau Yuen}
  \IEEEauthorblockA{
    School of Electrical and Electronics Engineering\\
    Nanyang Technological University\\
     Email: chau.yuen@ntu.edu.sg
  }

}

\begin{document}


\twocolumn
\maketitle
\begin{abstract}
Classification of multi-dimensional time series from real-world systems require fine-grained learning of complex features such as cross-dimensional dependencies and intra-class variations---all under the practical challenge of low training data availability. However, standard deep learning (DL) struggles to learn generalizable features in low-data environments due to model overfitting. We propose a versatile yet data-efficient framework, \textit{Intelligently Augmented Contrastive Tensor Factorization} (ITA-CTF), to learn effective representations from multi-dimensional time series. The CTF module learns core explanatory components of the time series (e.g., sensor factors, temporal factors), and importantly, their joint dependencies. Notably, unlike standard tensor factorization (TF), the CTF module incorporates a new contrastive loss optimization to induce similarity learning and class-awareness into the learnt representations for better classification performance. To strengthen this contrastive learning, the preceding ITA module generates targeted but informative augmentations that highlight realistic intra-class patterns in the original data, while preserving class-wise properties. This is achieved by dynamically sampling a “soft” class prototype to guide the warping of each query data sample, which results in an augmentation that is intelligently pattern-mixed between the “soft” class prototype and the query sample. These augmentations enable the CTF module to recognize complex intra-class variations despite the limited original training data, and seek out invariant class-wise properties for accurate classification performance. The proposed method is comprehensively evaluated on five different classification tasks, including extremely challenging tasks such as fault localization with more than 50 classes. Compared to standard TF and several DL benchmarks, notable performance improvements up to 18.7\% were achieved. 

\end{abstract}

\begin{IEEEkeywords}
Multi-dimensional time series, similarity matching, contrastive learning, dynamic data augmentation, fine-grained classification.
\end{IEEEkeywords}

\section{Introduction}
\label{intro}
Multi-dimensional time series are constantly being generated from the interconnected network of sensors found in critical real-world systems, such as industrial plants \cite{ren2023deep, ouyang2022new}, electric grids \cite{li2023dynamic, zheng2022multi, khan2023comparing}, transportation networks \cite{li2023dynamic, cuturi2011fast}, and wireless communication systems \cite{zhang2023variance, zhang2020integrated}. Bolstered by the remarkable success in computer vision and natural language processing domains, deep learning (DL) methods have been increasingly adopted to tackle a multitude of high-impact classification problems involving sensor time series. These problems can range from plant-level fault diagnosis \cite{zheng2022multi} to outdoor terrain classification for mobile autonomous systems \cite{ahmadi2020qcat}. Although notable advancements have been made in the time series classification field with DL methods  \cite{ouyang2022new, hao2023micos, bianchi2020reservoir,karim2019multivariate}, their reliance on large, labeled training data makes it challenging to extend supervised DL approaches to real-world applications, which are often subject to labeled data scarcity and data complexity \cite{adadi2021survey, zhang2020time}.

Labeled datasets for real-world applications are often scarce in practice as the time series data produced by complex systems typically do not have easily recognizable patterns and thus, require costlier expert knowledge for data labeling. Consequently, far less data is labeled than the amount generated. For instance, only 2\% of the disturbance events in power grids are analyzed and hand-labeled by utility personnel due to time and manpower constraints \cite{wilson2020automated}. Given the challenging combination of high-dimensional sensor data and few labeled training samples, supervised DL models may struggle to learn generalizable representations and risk overfitting to the limited data \cite{leon2020deep, arunan2023federated}. 

Time series data from real-world systems are also complex in several ways, which challenges the learning of useful feature representations for classification tasks:
\begin{enumerate}[1)]
\vspace{-0.1cm}
  \item In multi-dimensional time series from a network of sensors,  discriminatory features that aid class separation may be present not just in the autocorrelations of individual series, but also in the cross-dimensional interactions, e.g., between different sensor variables or between sensor and temporal factors \cite{ruiz2021great, zhang2020integrated}. However, conventional DL methods do not explicitly model these cross-dimensional interactions. For instance, in multi-channel deep convolution neural networks (MC-DCNN) \cite{zheng2014time}, each channel focuses on a single variable of the multi-dimensional time series and cross-channel dependencies are not explicitly considered.
\vspace{-0.1cm}
  \item Due to possible correlations of feature variables in the multi-dimensional time series, a natural idea is often to identify a smaller subset of features that are most relevant for the classification task. However, naive feature selection and dimensionality reduction methods generally do not account for the coupled nature of sensors in interconnected systems \cite{dey2019tensor}, resulting in the loss of joint information on cross-dimensional interactions that are crucial for accurate classification.
\vspace{-0.1cm} 
  \item Classification problems in real-world systems require more \textit{fine-grained} classification capabilities. This is due to the fact that datasets generated under varying operating conditions exhibit a considerable degree of intra-class variability within the same class, and yet, they may also be required to be classified into target classes that are similar to each other. For example, training data available in practice for outdoor terrain classification models may exhibit significant intra-class variations when collected from robots under different operating conditions (e.g., walking speeds, step frequencies, etc.). Despite the training data complexity, outdoor terrain classification models still need to finely learn between highly similar yet distinct terrains (e.g., dirt vs. gravel) in order for robots to quickly adapt to the challenges of the terrain \cite{ahmadi2020qcat}.
\end{enumerate}
Among existing studies that aim to learn cross-dimensional interactions in multi-dimensional time series, many focus primarily on spatial cross-channel dependencies. For instance, Karim \textit{et al.} \cite{karim2019multivariate} propose a hybrid convolutional neural network (CNN) module with a squeeze-and-excitation block to learn inter-dependencies across channels. Meanwhile, Zhang \textit{et al.} \cite{zhang2020tapnet} perform random permutations of channels prior to their attention prototypical network to induce learning of cross-channel interactions. However, these approaches are not designed to preserve or capture the inherent structural dependencies in the data across both spatial and temporal dimensions, which are crucial for modeling interconnected industrial systems. Consequently, while these approaches achieve competitive performance on benchmark datasets, their effectiveness deteriorates in real-world time series \cite{zheng2022multi}, which often have limited training data and yet high complexity, due to correlated feature dimensions and diverse target classes. Thus, there is a growing need for more well-designed and effective representation learning techniques that bolster multi-dimensional time series classification.

One promising technique for representation learning of multi-dimensional data is tensor factorization (TF) \cite{zhou2016linked}. In essence, TF is a feature extraction technique
that produces a structurally faithful decomposition of the raw tensor data into its key component factors (e.g., sensor factors, temporal factors) and a coefficient vector that captures the joint interactions of the factors \cite{zhou2016linked}. Moreover, due to their dimensionality reduction property, TF models are typically less parameter-heavy than DL models \cite{liu2023tensor}. Thus, their lower model complexity, combined with an inherent ability to capture important structural dependencies, make TF models well-suited for mitigating model overfitting challenges that arise when learning from multi-dimensional data with limited training samples.

However, as standard TF models are focused on reducing reconstruction losses, they are not inherently designed to learn class-aware representations or minimize classification errors in classification tasks \cite{qiu2021semi}. This means that these models cannot ensure a crucial requirement for good class separability--which is that the features extracted for data samples of the same class are similar to one another and dissimilar from extracted features of a different class. This shortcoming becomes especially pertinent under low-data settings, where highly efficient and fine-grained learning of discriminative features is needed from the limited training data available. 

Thus, to instill class-awareness into the TF-based feature extractor under data scarcity, we harness contrastive learning (CL), an unsupervised representation learning technique, known for its outstanding instance discrimination and similarity alignment properties. During CL, data samples and their noisy augmentations are contrasted against each other to learn  discriminative features that can distinguish similar samples of the same class from dissimilar samples of another class \cite{chen2020simple}. Through well-designed choice of data augmentations and contrastive loss functions, CL has  been widely successful in learning informative yet generalizable representations for a myriad of vision-related tasks \cite{xiao2021what, xie2021detco, zhao2021contrastive}. 

However, CL for time series is still in its infancy as it is challenging to design general data augmentation strategies that respect temporal dependencies while still being able to induce the learning of deeper underlying invariant properties of the time series \cite{zhang2022self}. Thus, existing works often use either general (e.g., jittering, permutations) \cite{ijcai2021-324} or domain knowledge-advised augmentations (e.g., band-pass filtering, 3D position rotation) \cite{yang2022atd}. However, these perturbations do not guarantee that the resultant augmentations still adhere to the general characteristics of a class, i.e., remain class-aware, which is crucial for classification tasks. Moreover, the spatio-temporal nature of multi-dimensional, networked time series
further complicates the development of a universally effective data augmentation strategy. Coupled with the challenge of training data scarcity, data augmentations need to be exceptionally targeted and efficient in highlighting relevant patterns for the feature extractor. Thus, we are motivated to seek a versatile and task-agnostic data augmentation strategy that makes efficient use of the information present in the limited training data. Specifically, the augmentations should be informative enough to teach a feature extractor to be explicitly aware of both intra- and inter-class patterns in low-data settings.

In this work, we address the shortcomings of two separately promising strategies (i.e., lack of class-awareness for TF and effective data augmentation strategies for time series CL), and propose a novel, generalizable approach for tackling classification problems involving multi-dimensional, networked time series in low-data settings. The proposed framework consists of three simple, connected modules:
\hspace{0.01cm}\raisebox{.5pt}{\textcircled{\raisebox{-.9pt}{1}}} an intelligently targeted augmentation (ITA) module, 
\hspace{0.01cm} \raisebox{.5pt}{\textcircled{\raisebox{-.9pt}{2}}} a class-aware contrastive tensor factorization (CTF)-based feature extractor that is also cognizant of important intra-class variations via the intelligent augmentations, and \hspace{0.01cm} \raisebox{.5pt}{\textcircled{\raisebox{-.9pt}{3}}} a downstream multi-layer perceptron (MLP) classifier.

As CL's representation learning performance is closely tied to the quality of augmentations, the ITA module aims to generate targeted and informative augmentations that contain intra-class patterns that are realistic to the dataset, while still preserving general class-wise characteristics. This is achieved by dynamically determining a ``soft'' class prototype within a mini-batch of samples, which serves as a reference sample for guiding the dynamic time warping (DTW) of each query sample. The resultant warped version (augmentation) of the query sample is thus a meaningful blend of the ``soft'' class prototype and the query sample, mimicking potential intra-class variations. The original samples alongside these intelligently pattern-mixed augmentations are fed to the CTF feature extractor to refine its ability to recognize complex intra-class patterns despite the limited number of original training data. The CTF module leverages CL’s similarity alignment property to compare and contrast samples, learning to ignore irrelevant intra-class variations and seek out invariant properties that encourage the coefficient vectors from tensor factorization to be closely aligned for samples of the same class. Combining the feature extractor with the third module, a MLP classifier, yields the final classification.

On the whole, the main contributions of this paper are:
\begin{enumerate}[1)]
\vspace{-0.15cm}
  \item We propose a versatile and data-efficient approach for learning effective representations from multi-dimensional time series with cross-factor interactions, under practical challenges of low training data and high intra-class variability.
  \vspace{-0.1cm}
  \item We incorporate into the CL process, a task-agnostic but data-guided augmentation strategy to capture realistic intra-classs variations that are inherent to the dataset. This enhances the CTF feature extractor's ability to recognize both intra- and inter-class variations despite the limited training data.
  \vspace{-0.1cm}
  \item We demonstrate the robust effectiveness of our proposed intelligently augmented CTF framework on three high-impact domain applications, spanning
five different classification tasks---including extremely challenging tasks such as fault localization with more than 50 classes.

\end{enumerate}

The remainder of the paper is organized as follows.
Section \ref{background} briefly covers the theoretical foundations of tensor factorization, contrastive learning, and dynamic time warping. Section \ref{proposed_method} formulates the proposed ITA-CTF framework. Section \ref{experiments} discusses the datasets, experiments, and results. Section \ref{conclusion} summarizes and concludes the work.

\section{Preliminary}
\label{background}
To better understand the proposed ITA-CTF framework, this section discusses
some key background on tensor factorization, contrastive learning, and dynamic time warping, together with the gaps that need to be filled for the respective methods.

\subsection{Tensor Factorization}
Tensors, i.e., multi-way arrays, are an effective means of representing relational attributes in multi-dimensional data affected by multiple factors \cite{kolda2009tensor}.  For instance, a multi-sensor time series dataset can be regarded as a third-order tensor with three modes, \textit{samples} $\times$ \textit{sensors} $\times$ \textit{time}. Within the dataset, each sample is then a second-order tensor, which is also commonly known as a matrix.

Tensor factorization (TF) models seek to capture  compact yet informative representations of high-dimensional tensors by decomposing the tensor into its component factors (e.g., sensor factors, temporal factors, etc.) and the learnable weight coefficients, which quantify the contributions of the factors and their joint interactions. Canonical polyadic decomposition (CPD) \cite{harshman1970foundations, carroll1970analysis} and Tucker decomposition \cite{tucker1966some, kolda2009tensor} are two of the most widely used TF models for learning cross-factor interactions in multi-dimensional data. However, Tucker decomposition is generally non-unique, while CPD provides unique solutions under mild conditions \cite{harshman1970foundations}, ensuring that it captures true latent factors instead of arbitrarily transformed ones \cite{papalexakis2016tensors}. Moreover, compared to Tucker decomposition, CPD has lower complexity (fewer model parameters) \cite{fernandes2021tensor} and offers greater interpretability \cite{papalexakis2016tensors}. Thus, we focus on CPD for the subsequent discussion. 

\begin{figure}[h!]
\centering
\includegraphics[width=1\linewidth]{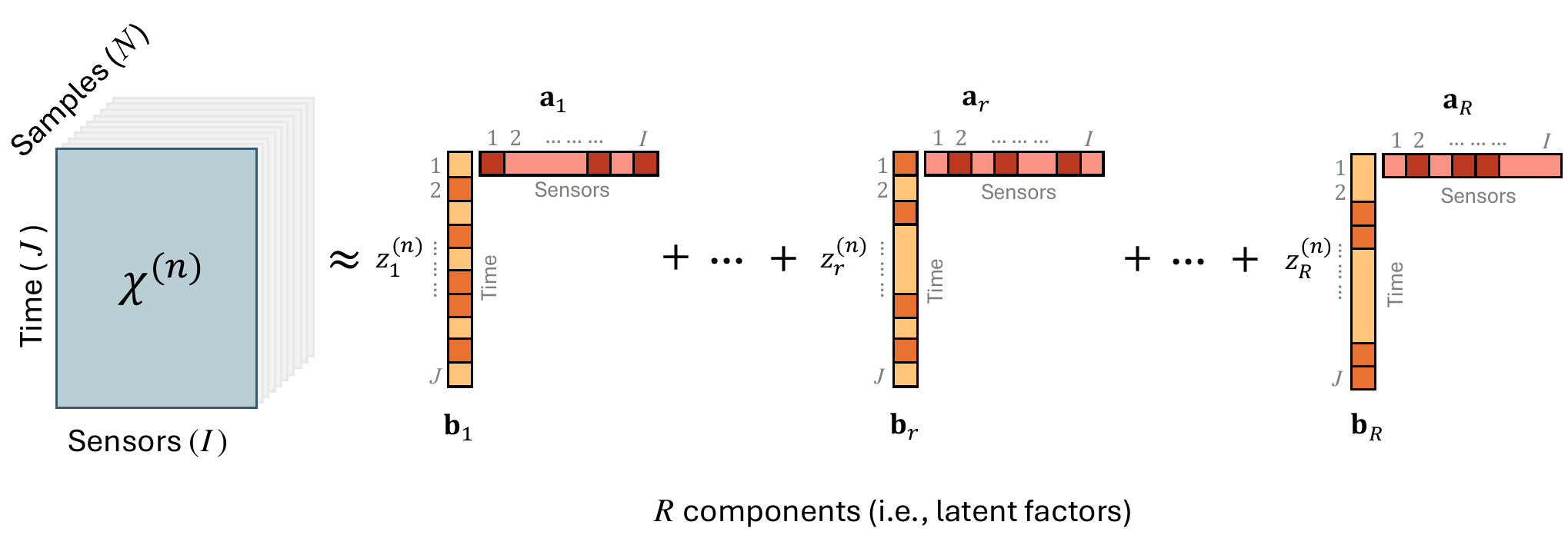}
\vspace{-1em}
\small\caption{Illustration of tensor factorization applied to a slice (n-th sample) of multi-dimensional tensor data, decomposing it into $R$ components (latent factors). Darker areas in vectors $\mathbf{a}_r$ and $\mathbf{b}_r$ represent higher contributions from sensors or time steps to the respective latent component. The weight coefficients $z_r$ quantify the importance of each component.}
\label{TF_explanation}
\end{figure}

Without loss of generality, we consider a third-order tensor of multi-sensor time series,
$$
\mathcal{X}=\left[\mathcal{X}^{(1)}, \mathcal{X}^{(2)}, \ldots, \mathcal{X}^{(N)}\right] \in \mathbb{R}^{N \times I \times J} ,
$$
where $\mathcal{X}^{(n)} \in \mathbb{R}^{I \times J}$ denotes the $n$-th sample; $N$, $I$, and $J$ are, respectively, the total samples, sensor channels, and time. In an effort to learn a faithful low-dimensional representation of the data, CPD aims to reconstruct the original noisy $\mathcal{X}^{(n)}$ by a low-rank approximation $\mathcal{\hat{X}}^{(n)}$ given by the weighted sum of rank-one tensor components $\upvartheta_r$ and the weight coefficients $z_r^{(n)} $ :
\begin{equation}
\begin{aligned}
\mathcal{\hat{X}}^{(n)} &= \sum_{r=1}^R z_r^{(n)} \cdot \upvartheta_r, \hspace{0.2cm} \text{where} \hspace{0.2cm}
\upvartheta_r = \mathbf{a}_r \circ \mathbf{b}_r, \quad r \in \{1, \ldots, R\} \\
&= z_1^{(n)}\mathbf{a}_1 \circ \mathbf{b}_1 + z_2^{(n)}\mathbf{a}_2 \circ \mathbf{b}_2 + \ldots +  z_R^{(n)}\mathbf{a}_R \circ \mathbf{b}_R
\end{aligned}
\end{equation}
where vectors $\mathbf{a}_r$ and $\mathbf{b}_r$ are learnable parameters, whose outer product yields $\upvartheta_r$; $R$ is the number components required for the original $\mathcal{X}^{(n)}$ to be well approximated. Intuitively, each r-th component $\upvartheta_r$ represents an underlying latent factor that explains the variability in $\mathcal{X}^{(n)}$. In the spatial dimension, $\mathbf{a}_r$ captures the contribution of each sensor to this latent factor, while  in the temporal dimension, $\mathbf{b}_r$ characterizes the evolution of this factor over time (see Fig. \ref{TF_explanation}). The relative importance of each r-th component is reconstructing $\mathcal{X}^{(n)}$ is determined by the weight coefficients $z_r^{(n)}$. By computing a weighted sum over the $R$ components, the original $\mathcal{X}^{(n)}$ can be reconstructed in a structured and interpretable manner, while preserving joint spatio-temporal interactions. Typically, $\mathbf{a}_r$ and $\mathbf{b}_r$ are accumulated into matrices, $
\mathbf{A}=\left[\mathbf{a}_{1}, \ldots, \mathbf{a}_{R}\right] \in \mathbb{R}^{I \times R}$ and $
\mathbf{B}=\left[\mathbf{b}_{1}, \ldots, \mathbf{b}_{R}\right] \in \mathbb{R}^{J \times R}$, called factor or basis matrices \cite{zhou2016linked}. The weights $z_r^{(n)} $ are accumulated into the coefficient vector $\mathbf{z}^{(n)} = \left[z_{1}^{(n)}, \ldots, z_{R}^{(n)}\right] \in \mathbb{R}^{1 \times R}$.

The goal of CPD is to learn high-quality factor matrices and weight coefficients that minimize the reconstruction loss $\mathcal{L}_{rec}$ between the original data and its low-rank approximation\footnote{In general, for a $v$-th order input tensor, there will be $v$ factor matrices in Equation 2.}:
\begin{equation}
\label{reconstruction_loss}
\begin{split}
\underset{\mathbf{Z}, \mathbf{A}, \mathbf{B}}{\arg \min}\hspace{0.5em} \mathcal{L}_{rec} &= \sum_{n=1}^N\left\|\mathcal{X}^{(n)}-\llbracket \mathbf{z}^{(n)}, \mathbf{A}, \mathbf{B} \rrbracket\right\|_F^2 \\
&= \|\mathcal{X}-\llbracket \mathbf{Z}, \mathbf{A}, \mathbf{B} \rrbracket\|_F^2,
\end{split}
\end{equation}
where 
$\llbracket.\rrbracket$ is the compact notation for the sum of outer products and $\|.\|_F$ is the Frobenius norm. As seen from Eq. \ref{reconstruction_loss}, while standard CPD is well-suited for capturing multi-factor relationships that aid faithful data reconstruction, it is not inherently designed to learn class-aware representations or minimize downstream classification errors. 

\subsection{Contrastive Representation Learning}

\begin{figure}[h]
\centering
\includegraphics[width=0.8\linewidth]{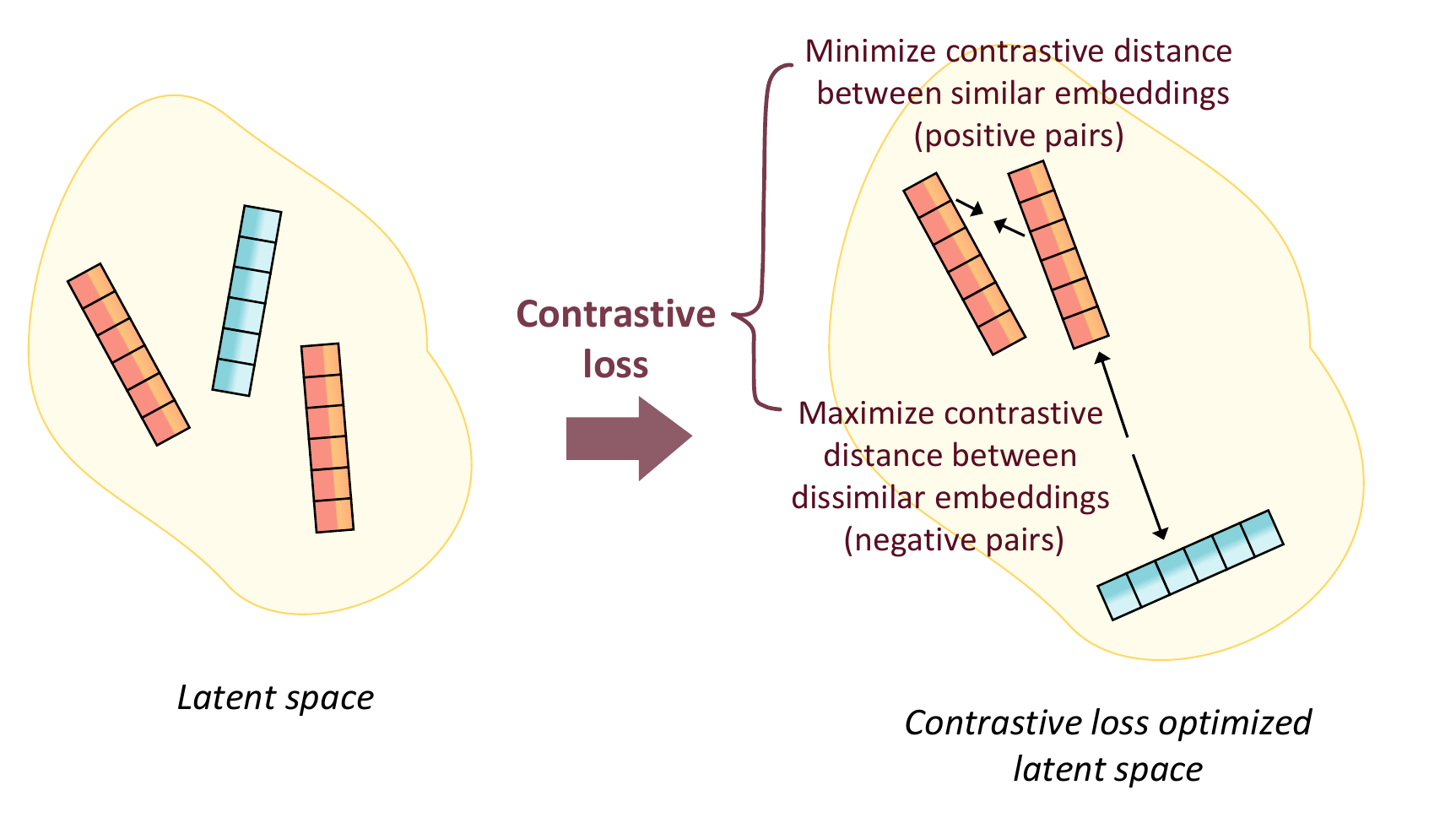}
\vspace{-1em}
\small \caption{Contrastive learning, where embeddings of similar samples are pulled closer together and embeddings of dissimilar samples are pushed further apart.}
\label{CL_explanation}
\end{figure}

Contrastive learning (CL) is a powerful representation learning technique, where data samples are contrasted against each other to learn  discriminative features that can distinguish similar samples of the same distribution from dissimilar samples of another distribution \cite{chen2020simple}\cite{khosla2020supervised}. A contrastive loss function is optimized to learn representations in a latent space, where the distance between  similar (positive) embedding pairs is minimized, and the distance between dissimilar (negative) embedding pairs is maximized, as shown in Fig. \ref{CL_explanation}. Optimizing the contrastive loss, thus, inherently promotes learning of semantically related, class-aware features that explain the distance variations of the embeddings. 

A key factor influencing the quality and generalizability of the learnt representations from CL is the choice of positive and negative pairs \cite{xiao2021what}. A positive pair is a data sample and a noisy augmentation of that sample, while a negative pair can be a sample from another distribution or--for a deeper learning of invaraint representations--its augmentation. Thus, a well-designed choice of negative pairs and data augmentations is crucial for CL to learn high-quality representations amidst labeled data scarcity.

However, for time series, it is challenging to design general yet impactful data augmentation strategies that can induce the learning of deeper underlying invariant properties of the time series while still respecting temporal dependencies \cite{zhang2022self}. Thus, augmentation strategies proposed in existing works are usually tailored to the applications studied. For example, Zhang \textit{et al.}\cite{zhang2022self} create augmentations through domain knowledge-advised perturbations of frequency components in frequency-sensitive time series. Meanwhile, Eldele \textit{et al.}\cite{ijcai2021-324} employ a considerably general augmentation strategy of noise jittering and time series segment permutation. However, the permutation strategy risks disrupting temporal dependencies and creating unrealistic patterns for the data distribution.

\subsection{Dynamic Time Warping}
\label{DTW_background_section}
Here, we briefly recap the basics of dynamic time warping (DTW) \cite{sakoe1978dynamic} as its pattern mixing and time warping properties form the inspiration for the proposed data augmentation strategy. 
DTW is a classic similarity matching algorithm that can synchronize two time series with different temporal dynamics (e.g., different speeds, lags) by optimally aligning their similar patterns on a common time axis.

Assume we have a query sample $\mathcal{X}^{(n)} = \left[\mathbf{x}_{1},...,\mathbf{x}_{k},..., \mathbf{x}_{K}\right]^{\top}$ and a reference sample $\mathcal{X}^{(n^{\prime})} = \left[\mathbf{x^{\prime}}_{1},...,\mathbf{x^{\prime}}_{l},..., \mathbf{x^{\prime}}_{L}\right]^{\top}$ measured from the same system, where $\mathbf{x}_{k}$, $\mathbf{x^{\prime}}_{l}$ are multi-sensor feature vectors at time index $k$ and $l$ respectively. DTW matches similar patterns between the two samples by calculating the lowest-cost warping path that matches data points in $\mathcal{X}^{(n)}$ to the most similar counterparts in $\mathcal{X}^{(n^{\prime})}$, using the time axis of $\mathcal{X}^{(n^{\prime})}$ as the reference scale. Here, warping refers to the stretching or compressing feature vectors of $\mathcal{X}^{(n)}$ along the time axis in order to match the temporal pace of $\mathcal{X}^{(n^{\prime})}$. Data points of the two series are considered optimally matched when the overall matching cost, defined as the total distance between all matched pairs, is minimized. 

With the optimal warping path, $\mathcal{X}^{(n)}$ can be warped to generate a new variant $\mathcal{\tilde{X}}^{(n)}$, which is a combination of the feature patterns from the original $\mathcal{X}^{(n)}$ and the reference sample $\mathcal{X}^{(n^{\prime})}$. Thus, if we aim to create augmentations that capture realistic intra-class variations while maintaining  general class-wise characteristics, the key then lies in intelligently selecting effective reference samples for warping the query samples.

\begin{figure*}[!t]
\centering
\includegraphics[width=1.0\textwidth]{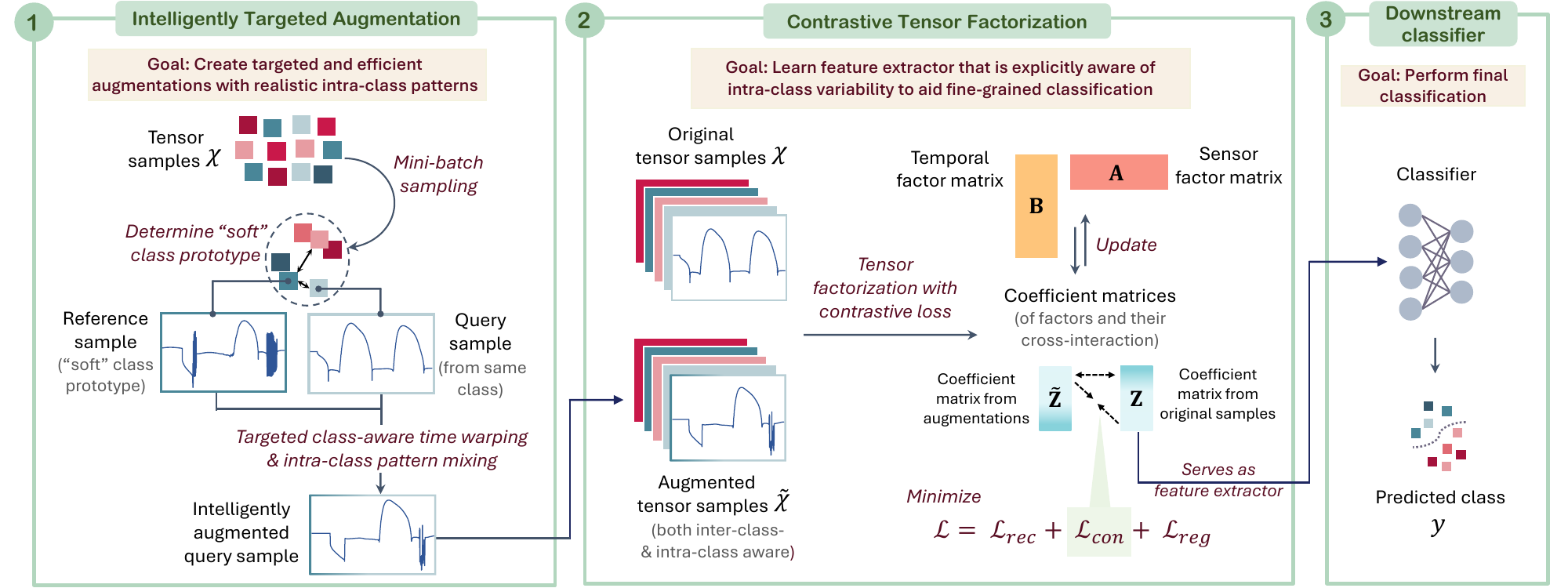}
\small \caption{The proposed framework consists of three simple but highly effective modules:
\raisebox{.5pt}{\textcircled{\raisebox{-.9pt}{1}}} an intelligently targeted augmentation  module, \raisebox{.5pt}{\textcircled{\raisebox{-.9pt}{2}}} a class-aware contrastive tensor factorization-based feature extractor that is also cognizant of important intra-class variations via the intelligent augmentations, and 
\raisebox{.5pt}{\textcircled{\raisebox{-.9pt}{3}}} a downstream MLP classifier.}
\vspace{-0.5 em}
\label{fig_proposed_framework}
\end{figure*}

\section{Contrastive Tensor Factorization with Intelligently Targeted Augmentation}
\label{proposed_method}

In this section, we present our proposed framework to tackle classification problems involving multi-dimensional, networked time series in low-data environments. The framework consists of three connected modules:
\raisebox{.5pt}{\textcircled{\raisebox{-.9pt}{1}}} an intelligently targeted augmentation (ITA) module, \raisebox{.5pt}{\textcircled{\raisebox{-.9pt}{2}}} a class-aware contrastive tensor factorization (CTF)-based feature extractor that is also cognizant of important intra-class variations through the intelligent augmentations, and 
\raisebox{.5pt}{\textcircled{\raisebox{-.9pt}{3}}} a downstream multi-layer perceptron (MLP) classifier.

Recalling that the effectiveness of representations learned during contrastive learning depends on the quality of augmentations, the ITA module aims to generate targeted and informative augmentations that explicitly highlight realistic intra-class patterns of the dataset, while preserving general class-wise characteristics. This is achieved by dynamically determining a ``soft'' class prototype within a mini-batch of samples, which serves as a reference sample for guiding the DTW-based warping of each query sample. The resultant warped version (augmentation) of the query sample is thus a meaningful blend of the ``soft'' class prototype and the query sample, mimicking potential intra-class variations. The original samples alongside the intelligent augmentations are fed into the CTF feature extractor to refine its ability to recognize complex intra-class patterns despite the limited number of original training data. The CTF module leverages CL’s similarity alignment property to compare and contrast samples, learning to ignore irrelevant intra-class variations and seek out invariant properties that encourage the coefficient vectors from tensor factorization to be closely aligned for samples of the same class. Combining the feature extractor with the third module, a downstream MLP classifier, yields the final classification for a data sample. The overall framework is visualized in Fig. \ref{fig_proposed_framework}, with details discussed henceforth.  

\subsection{Classification Problem Formulation}
Given training samples $\left\{ (\mathcal{X}^{(n)}, y^{(n)})\right\}_{n=1}^N$ from $P$ classes, we aim to train a model $ \mathcal{H}:\mathcal{X}^{(i)} \rightarrow y^{(i)}$ that accurately predicts the class label $y^{(i)} \in \lbrace 1,..., p,..., P  \rbrace$ of a new test sample $\mathcal{X}^{(i)}$. Here, model $ \mathcal{H}$ encompasses our enhanced ITA-CTF feature extractor and a MLP downstream classifier, which are trained on the training samples and their augmentations.

\subsection{ITA-CTF Feature Extractor}
With the key motivation of improving representation learning of multi-dimensional, networked time series in low-data settings, we formulate below the ITA-CTF feature extractor of our proposed framework.

\subsubsection{Intelligently Targeted Augmentation}
\label{proposed_dtw_aug_section}
To aid class-aware representation learning in CL, data augmentations need to preserve the class characteristics of the original samples. However, the choice of a universal augmentation strategy that ensures class preservation, while meaningfully capturing data variations is far from obvious. In this section, we present a task-agnostic but intelligently targeted data augmentation strategy, which preserves class labels while accounting for the intra-class variability in real-world data. The goal is to create a rich set of augmented samples $\mathcal{\tilde{X}}$ highlighting the intra-class variations from the original limited training samples $ \mathcal{X}$, thus maximizing overall data-efficiency.

The basic premise is that we can leverage DTW's pattern mixing and time warping properties (as described in Section \ref{DTW_background_section}) to create a warped version of any query sample using a reference sample as a guide. However, if the reference sample is chosen randomly, we cannot ensure that the warped (augmented) version of the query sample maintains general class characteristics, nor can we expect realistic intra-class patterns to be captured in the augmentation. 

Therefore, we utilize instead a ``soft'' class prototype as the reference sample to guide the DTW process and form a basis for the augmentation. A class prototype can be defined as the most archetypal or representative sample of the class. Mathematically, the class prototype is the sample with the maximal difference in distance between the centroid of samples from the same class and the centroid of samples from other classes. However, determining the single most protypical sample among all samples of the class is a time consuming process and overly deterministic. Instead, we dynamically determine a ``soft'' class prototype for each query sample by randomly sampling (with replacement) a mini-batch of samples of size $S \ll N$. The advantage of using a flexible ``soft'' class prototype to guide the augmentation process is that allows different intra-class variations to be captured while still constraining the augmentations to general class characteristics.

To calculate the difference in class-wise centroid distances $\Gamma(\cdot)$ for each sample $\mathcal{X}^{(s)}$ of the mini-batch, we utilize a simple yet effective nearest-centroid distance measure, recommended in \cite{iwana2017efficient, iwana2021time}, as it does not require extensive training like neural networks:
\begin{equation}
\begin{aligned}
\Gamma(\mathcal{X}^{(s)}) & = \frac{1}{\sum_{s^{\prime}}\left[p_{s^{\prime}} \neq p_s\right]} \sum_{s^{\prime}} \updelta(\mathcal{X}^{(s^{\prime})}, \mathcal{X}^{(s)}) \mid\left[p_{s^{\prime}} \neq p_s\right] \\
& -\frac{1}{\sum_{s^{\prime}}\left[p_{s^{\prime}}=p_s\right]} \sum_{s^{\prime}} \updelta(\mathcal{X}^{(s^{\prime})}, \mathcal{X}^{(s)}) \mid\left[p_{s^{\prime}}=p_s\right],
\end{aligned}
\end{equation}
where $\updelta(\cdot)$ is the DTW distance, the top term is the average centroid distance between the current sample $\mathcal{X}^{(s)}$ with class label $p_s$ and all other samples $\mathcal{X}^{(s^\prime)}$ with different labels, and the bottom term is average centroid distance between $\mathcal{X}^{(s)}$ and all other samples $\mathcal{X}^{(s^\prime)}$  with the same label.  The ``soft'' class prototype  $\mathcal{X}^{(s_{p}^{*})}$ is then the sample that yields the maximum difference in the centroid distances:
\begin{equation}
\mathcal{X}^{(s_{p}^{*})}=\underset{\{s=1, \ldots, S\}}{\operatorname{arg max}} \Gamma(\mathcal{X}^{(s)})
\label{class_prototype_eqn}
\end{equation}

With the dynamically determined ``soft'' class prototype as the reference sample, the query sample is warped using DTW to create its augmentation. This process of determining a ``soft'' class prototype (which varies for each query sample due to mini-batch sampling) and the subsequent generation of the DTW-based augmentation is repeated for all training samples.

\subsubsection{Contrastive Tensor Factorization}
Here, our goal is to instill class-awareness into the learnt coefficient matrix $\mathbf{Z}$ of the CPD process by introducing a new contrastive loss to facilitate contrastive learning within the tensor factorization process.

First, the original tensor samples $\mathcal{X}$ and the augmented samples $\tilde{\mathcal{X}}$ are concurrently fed into the CPD process to yield the corresponding coefficient matrices $\mathbf{Z}$ and $\tilde{\mathbf{Z}}$. Then, we use the row vectors of $\mathbf{Z}$ and $\tilde{\mathbf{Z}}$ (which represent the feature extraction of each sample) to calculate the contrastive loss, whose minimization constrains the learned features in $\mathbf{Z}$ to be more class-aware. We now discuss the formulation of the contrastive loss function and its integration into CPD.

Any proposed contrastive loss function incorporated within CPD (or, generally, any other TF model) has to be amenable to the quadratic optimization that occurs during tensor factorization \cite{yang2022atd}. Thus, instead of the standard logistic- or softmax-like formulations such as noise contrastive estimation (NCE) \cite{gutmann2012noise} or normalised-temperature cross-entropy (NT-Xent)\cite{chen2020simple}, we consider a subtraction-based formulation for the contrastive loss $\mathcal{L}_{con}$:
\begin{equation}
\label{main_conloss}
\begin{aligned}
\mathcal{L}_{con} = & (\gamma+1) \cdot \frac{1}{N(N-1)} \sum_{n=1}^N \sum_{m \neq n}^N \left\langle\frac{\mathbf{z}^{(n)}}{\left\|\mathbf{z}^{(n)}\right\|_2}, \frac{\tilde{\mathbf{z}}^{(m)}}{\left\|\tilde{\mathbf{z}}^{(m)}\right\|_2}\right\rangle \\
& -\frac{1}{N} \sum_{n=1}^N \left\langle\frac{\mathbf{z}^{(n)}}{\left\|\mathbf{z}^{(n)}\right\|_2}, \frac{\tilde{\mathbf{z}}^{(n)}}{\left\|\tilde{\mathbf{z}}^{(n)}\right\|_2}\right\rangle,
\end{aligned}
\end{equation}
where the top term is the average cosine similarity of all possible pairs of non-corresponding row vectors of $\mathbf{Z}$ and $\tilde{\mathbf{Z}}$, and the bottom term is the average cosine similarity of all pairs of corresponding row vectors, i.e., positive pairs derived from the original sample and its augmentation; $\gamma$ is a tunable hyperparameter.  

Rewriting Eq. \ref{main_conloss} in a compact matrix notation for simplicity and brevity yields:
\begin{equation}
\label{matrix_conloss}
\begin{aligned}
\mathcal{L}_{con} = \operatorname{Tr}\left(\mathbf{Z}^{\top} \mathbf{D}(\mathbf{Z}) \mathbf{G}(\gamma) \mathbf{D}(\tilde{\mathbf{Z}}) \tilde{\mathbf{Z}}\right),
\end{aligned}
\end{equation}
where operator $\operatorname{Tr}(.)$ calculates the trace of a given matrix, 
$\mathbf{D}(\mathbf{Z})=\operatorname{diag}\left(1/{\left\|\mathbf{z}^{(1)}\right\|_2}, \cdots, 1/{\left\|\mathbf{z}^{(N)}\right\|_2}\right)$ is the row-wise normalization matrix, and $\mathbf{G}(\gamma)$ is a symmetric matrix absorbing the scalars, with diagonal entries being $-1/{N}$ and off-diagonal entries being $(\gamma+1)/{N(N-1)}$. 

The final loss for the CTF model, thus, considers the original reconstruction loss, the new contrastive loss, and a standard regularizer\cite{calvetti2003tikhonov} to prevent overfitting: 
$$\mathcal{L}=\mathcal{L}_{rec}+\beta \mathcal{L}_{con}+\mathcal{L}_{reg},$$
with
\begin{equation}
\begin{aligned}
\mathcal{L}_{rec} & = \|\mathcal{X} - \llbracket \mathbf{Z}, \mathbf{A}, \mathbf{B} \rrbracket \|_F^2 + \|\tilde{\mathcal{X}} - \llbracket \tilde{\mathbf{Z}}, \mathbf{A}, \mathbf{B} \rrbracket \|_F^2, \\
\mathcal{L}_{reg} &= \alpha \left( \|\mathbf{Z}\|_F^2 + \|\tilde{\mathbf{Z}}\|_F^2 + \|\mathbf{A}\|_F^2 + \|\mathbf{B}\|_F^2 \right)
\end{aligned}
\label{overall_loss}
\end{equation}
where $\alpha$ and $\beta$ are positive-valued tunable hyperparameters controlling the relative contribution of each type of loss.

This loss function $\mathcal{L}$ needs to minimized with respect to $\mathbf{A}$, $\mathbf{B}$, $\mathbf{Z}$, and $\tilde{\mathbf{Z}}$. However, as the loss function is non-convex with respect to $\mathbf{Z}$ and $\tilde{\mathbf{Z}}$, an iterative update-based version of the \textit{alternating least squares} (ALS) algorithm \cite{harshman1970foundations, carroll1970analysis} is employed to sequentially solve for $\mathbf{A}$, $\mathbf{B}$, $\tilde{\mathbf{Z}}$, and $\mathbf{Z}$.

Considering the subproblem with respect to $\mathbf{Z}$ , the goal is to obtain the optimal $\mathbf{Z}^*$ that minimizes Eq. \ref{overall_loss}, given  $\mathbf{A}$, $\mathbf{B}$, and $\tilde{\mathbf{Z}}$:
\begin{equation}
\label{optimalZ}
\begin{split}
\mathbf{Z}^* \leftarrow \underset{\mathbf{Z}}{\arg \min }(\|\mathcal{X}-\llbracket \mathbf{Z}, \mathbf{A}, \mathbf{B} \rrbracket\|_F^2+\alpha\|\mathbf{Z}\|_F^2 \\ + \hspace{0.1cm} \beta \operatorname{Tr}\left(\mathbf{Z}^{\top} \mathbf{D}(\mathbf{Z}) \mathbf{G}(\gamma) \mathbf{D}(\tilde{\mathbf{Z}}) \tilde{\mathbf{Z}}\right)) 
\end{split}
\end{equation}

To obtain the optimal $\mathbf{Z}^*$, the row-wise subproblems can be independently solved for all rows of $\mathbf{Z}$:

\begin{equation}
\begin{aligned}
\label{rowwise_optimization}
\underset{\mathbf{z}}{\arg \min} & \left(\left\|\mathcal{X}^{(n)} - \llbracket \mathbf{z}, \mathbf{A}, \mathbf{B} \rrbracket\right\|_F^2 + \alpha\|\mathbf{z}\|_F^2 \right. \\
& \left. + \beta \operatorname{Tr}\left(\frac{\mathbf{z}^{\top}}{\|\mathbf{z}\|_2} \mathbf{g}^{(n)} \mathbf{D}(\tilde{\mathbf{Z}}) \tilde{\mathbf{Z}}\right)\right),
\end{aligned}
\end{equation}
where $\mathcal{X}^{(n)}$ is the $n$-th sample of $\mathcal{X}$, and $\mathbf{g}^{(n)}$ is the $n$-th row of $\mathbf{G}(\gamma)$.

\begin{algorithm}[t]
\caption{\small Iterative ALS updating for CTF framework.}
\footnotesize
\algsetup{linenosize=\footnotesize}
\begin{algorithmic}[1]
\renewcommand{\algorithmicrequire}{\textbf{Input:}}
\renewcommand{\algorithmicensure}{\textbf{Output:}}
\newcommand{\algorithmicbreak}{\textbf{break}}
\newcommand{\BREAK}{\STATE \algorithmicbreak}
\REQUIRE Original tensor samples $\mathcal{X}$, initialized factors $\left\{\mathbf{A}^{1}, \mathbf{B}^{1}\right\}$, and \\ \hspace{3.6mm} corresponding augmented samples $\tilde{\mathcal{X}}$.
\ENSURE Optimized factors $\left\{\mathbf{A}^{Q}, \mathbf{B}^{Q}\right\}$ and coefficient matrix $\mathbf{Z}$ 
\\ \text{Initialize} epoch $q = 1$
\WHILE {$q \leq Q$ or change in loss $ < 0.1\%$ for 5 consecutive epochs}  
 \FOR {tensor batch $\mathcal{X}^{q}$ and augmentations $\tilde{\mathcal{X}_{q}}$}
 \STATE Given $\mathbf{A}^{q}$ and $\mathbf{B}^{q}$, initialize $\mathbf{Z}$ based on $\mathcal{X}^{q}$; 
 \STATE Given $\mathbf{A}^{q}$ and $\mathbf{B}^{q}$, initialize $\tilde{\mathbf{Z}}$ based on $\tilde{\mathcal{X}_{q}}$; 
 \STATE Given $\mathbf{A}^{q}$, $\mathbf{B}^{q}$, and $\tilde{\mathbf{Z}}$, update $\mathbf{Z}$ using Eq. \ref{z_improved}; 
 \STATE Given $\mathbf{A}^{q}$, $\mathbf{B}^{q}$, and $\mathbf{Z}$, update $\tilde{\mathbf{Z}}$ using Eq. \ref{z_improved};
 \STATE Given $\mathbf{B}^{q}$, $\mathbf{Z}$, and $\tilde{\mathbf{Z}}$, solve for $\mathbf{A}^{q+1}$ as a least squares problem;
 \STATE Given $\mathbf{A}^{q}$, $\mathbf{Z}$, and $\tilde{\mathbf{Z}}$, solve for $\mathbf{B}^{q+1}$ as a least squares problem;
 \ENDFOR
 \STATE $q=q+1$
\ENDWHILE
\end{algorithmic} 
\end{algorithm}

Taking the derivative of Eq. \ref{rowwise_optimization} and equating it to zero, gives the solution to the sample-/row-wise subproblem:
\begin{equation}
\label{z_solution}
\mathbf{z}=\mathbf{w}_1 \mathbf{W}_3-\frac{\beta \mathbf{w}_2}{2\|\mathbf{z}\|_2}\left(\mathbf{I}-\frac{\mathbf{z}^{\top} \mathbf{z}}{\|\mathbf{z}\|_2^2}\right) \mathbf{W}_3
\end{equation}
where $
\mathbf{w}_1=\mathbf{X}_1^{(n)}(\mathbf{A} \odot \mathbf{B})$, \quad $\mathbf{w}_2=\mathbf{g}^{(n)} \mathbf{D}(\tilde{\mathbf{Z}}) \tilde{\mathbf{Z}}$, \quad $\mathbf{W}_3=\left(\mathbf{A}^{\top} \mathbf{A} * \mathbf{B}^{\top} \mathbf{B}+\alpha \mathbf{I}\right)^{-1}$.
Here, $\mathbf{X}_1^{(n)}$ is the 1-mode unfolding of $\mathcal{X}^{(n)}, \odot$ is the column-wise Kronecker product,  
and $*$ is the Hadamard product. 

The iterative update-based approach is then to, for each $n$-th row of $\mathbf{Z}$, start with an initial guess for $\mathbf{z}_{in}$ and iteratively improve upon the solution $\mathbf{z}_{im}$ as follows:
\begin{equation}
\mathbf{z}_{im} =\mathbf{w}_1 \mathbf{W}_3-\frac{\beta \mathbf{w}_2}{2\|\mathbf{z}_{in}\|_2}\left(\mathbf{I}-\frac{\mathbf{z}_{in}^{\top} \mathbf{z}_{in}}{\|\mathbf{z}_{in}\|_2^2}\right) \mathbf{W}_3
\label{z_improved}
\end{equation}
where $\mathbf{z}_{in}$ is first obtained by solving the least squares problem when $\beta = 0$ in Eq. \ref{rowwise_optimization}. The resultant $\mathbf{z}_{im}$ will then become the $\mathbf{z}_{in}$ in the next round to iteratively improve the solution. Interested readers may also refer to, for example, \cite{yang2022atd}, for a convergence analysis, which is beyond the scope of this work. 

The complete process of minimizing Eq. \ref{overall_loss} by alternatively updating $\mathbf{A}$, $\mathbf{B}$, $\mathbf{Z}$, and $\tilde{\mathbf{Z}}$ is given in Algorithm 1. Once the feature extractor $\mathbf{Z}$ is sufficiently trained, it can be utilized to extract the features of  new tensor samples. These feature extractions can then be fed to a downstream classifier for the final classification, as discussed in the next section.

\section{Experiments and Results}
\label{experiments}
In this section, we introduce the datasets used, and provide a comprehensive performance analysis of the proposed method with comparative studies against benchmarks, ablation studies, and a sensitivity analysis of the key hyperparameters. 

\subsection{Dataset Description}
\label{datasets_subsection}
Our main research motivation is that, when dealing with multi-dimensional time series from  real-world systems with limited training data, more effective representation learning strategies are needed to extract complex features while avoiding model overfitting. To evaluate our model in realistic conditions, we carefully select three challenging multi-dimensional time series datasets that reflect operational complexities and offer only limited amounts of labeled training data. As seen in Table \ref{data_description}, the three datasets span a diverse range of high-impact domains---including autonomous robot navigation, traffic, and power grids---and vary in terms of the number of channels, sequence lengths, and classes. Moreover, the resulting classification tasks cover five quintessential problems with strong real-world applicability, including the highly challenging fault localization task involving more than 50 classes. Further details of the datasets are described below.

\begin{table}{}
\newcolumntype{P}[1]{>{\centering\arraybackslash}p{#1}}
\renewcommand{\arraystretch}{1.3}
\scriptsize
\centering
\caption{\centering Comparative overview of dataset attributes.}
\vspace{0.1cm}
\label{data_description}
\begin{tabular}{P{1.8cm}P{1.3cm}P{0.6cm}P{1cm}P{1.7cm}P{0.7cm}}
\hline Dataset \hspace{0.5cm}name & No. of train samples & Channels ($I$) & Time steps ($J$) & Classification task &  No. of classes \\
\hline \centering QCAT Robot \cite{ahmadi2020qcat} & 460       & 22   & 662  & Terrain type   & 6   \\
\centering PEMS Traffic \cite{cuturi2011fast}    & 267      & 963   & 144 & Day of the week   & 7  \\
\multirow{3}{1.7cm}{\centering PSML Electric Grid \cite{zheng2022multi} }    & \multirow{3}{*}{439}    & \multirow{3}{*}{91}   & \multirow{3}{*}{960} & Fault type & 5  \\
          &        &    &  & Fault location  & 53   \\
          &        &    &   & Fault start time & 174  \\
\hline
\end{tabular}
\end{table}

\subsubsection{QCAT Legged Robot Terrain Classification Dataset}
Accurate terrain classification enables robots to correctly, and automatically, adjust their configurations (e.g., stride length, footfall position) to overcome the specific challenges of different terrains. The dataset \cite{ahmadi2020qcat} consists of sensor measurements generated from an autonomous robot as it traverses different outdoor terrains near the Queensland Centre for Advanced Technologies (QCAT), Australia. There are 22 interconnected feature variables (e.g., linear acceleration, angular velocity, etc.) recorded from the robot's force sensors and inertial measurement units. There are 6 terrain types for classification: concrete, grass, gravel, mulch, dirt, and sand. 

\begin{figure}[!h]
\vspace{-1.2 em}
\centering
\subfloat[Sample with speed = 1]{\includegraphics[width=0.4\linewidth]{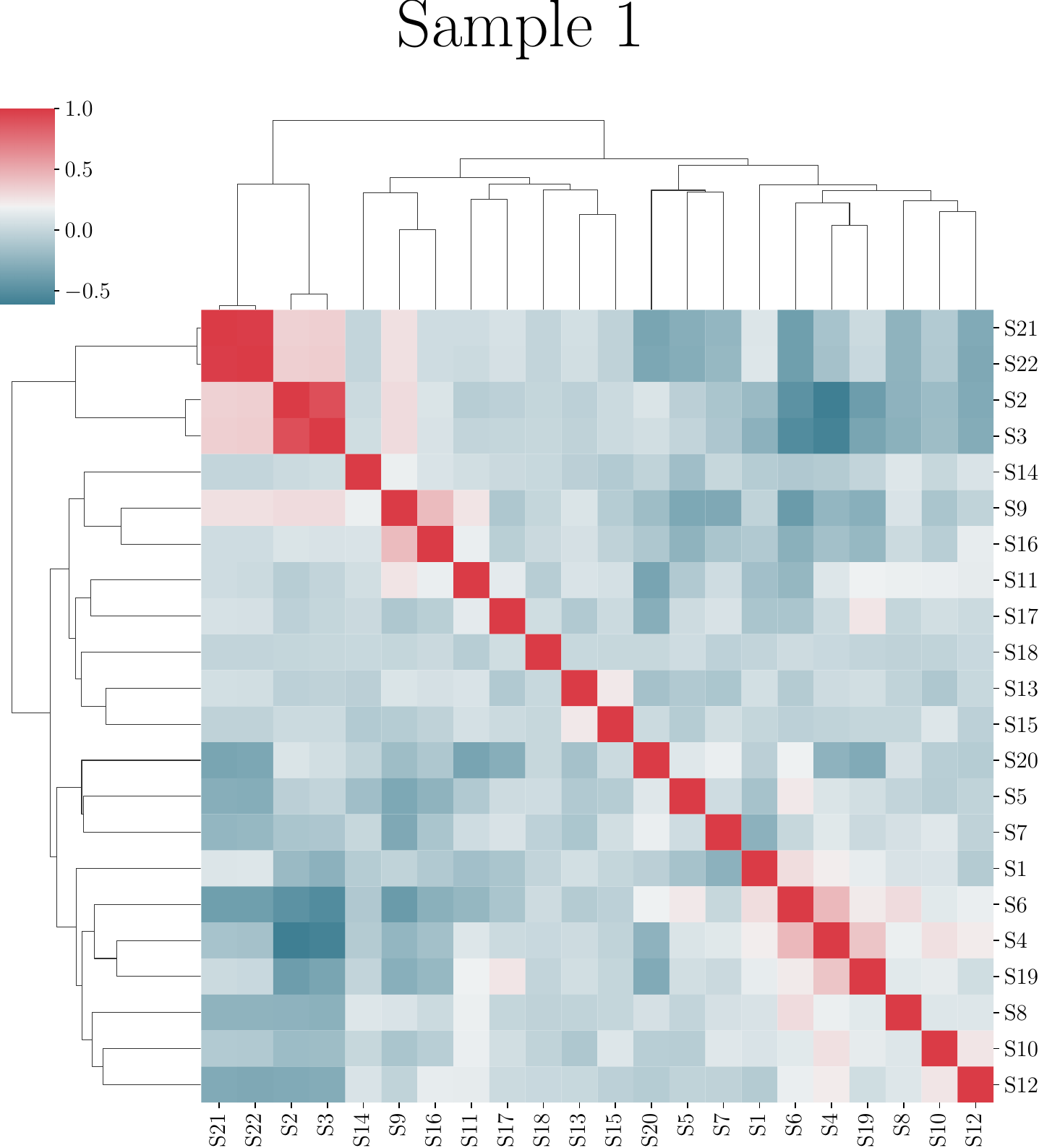}%
\label{qcat_speed1}}
\hfil 
\subfloat[Sample with speed = 6]{\includegraphics[width=0.4\linewidth]{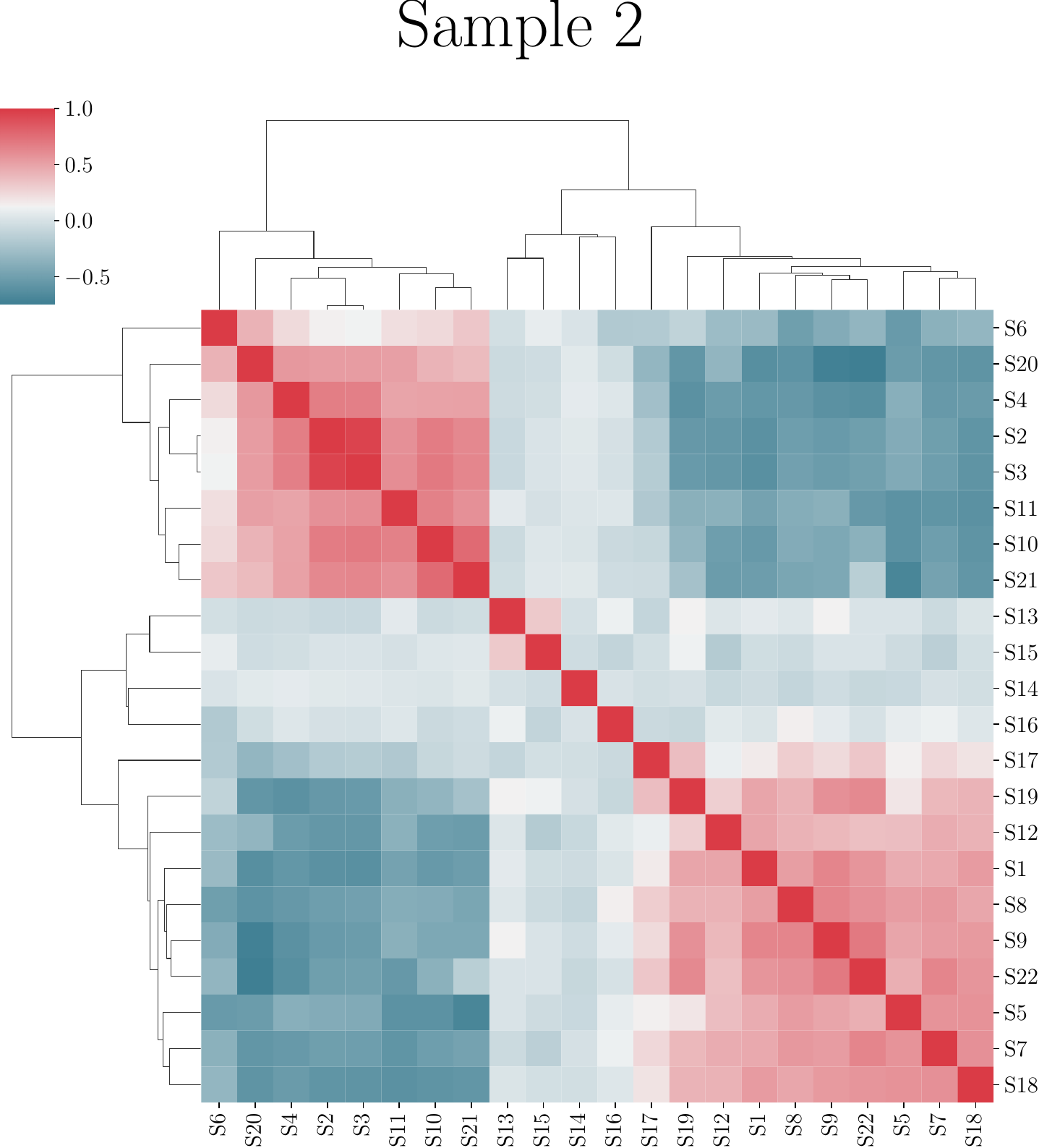}%
\label{qcat_speed6}}
\small \caption{Two random samples from same terrain class label can have considerable intra-class variations (as shown by the correlations of sensors) due to varied operating conditions such as different robot walking speeds.}
\label{qcat_fig}
\end{figure}

For each terrain type, the data was collected under various conditions such as 6 different robot walking speeds, 2 different step lengths, 3 different step frequencies, and so on. Thus, the data exhibits a considerable degree of intra-class variation (see Fig. \ref{qcat_fig} for instance). The original dataset contains 2,880 samples. To simulate a low-data scenario, we create a smaller yet representative experimental dataset by selecting every fifth sample from each terrain type, resulting in a dataset that is 20\% of the original. We then adopt a 80\%-20\% train-test split.

\subsubsection{PEMS Traffic Dataset}
The dataset \cite{cuturi2011fast} focuses on day-of-the-week classification based on traffic patterns, and it is available on the popular UCI Machine Learning Repository \cite{pems_sf}. The data contains car lane occupancy rates collected throughout the day, across a network of measurement stations in San Fransico, by the California Department of Transportation Performance Measurement System (PEMS). There are a staggering 963 stations (i.e., feature variables) recorded for each time series. The daily time series are collected over 15 months, thus, the dataset contains potential seasonal as well as weekday-weekend variations. As seen from Table \ref{data_description}, the original dataset is already small-sized and challenging, therefore, no further downsizing is needed. We adopt the dataset's pre-specified train and test set.

\subsubsection{PSML Electric Grid Dataset}
\begin{figure}[!t]
\vspace{0.5em}
\centering
\includegraphics[width=0.6\linewidth]{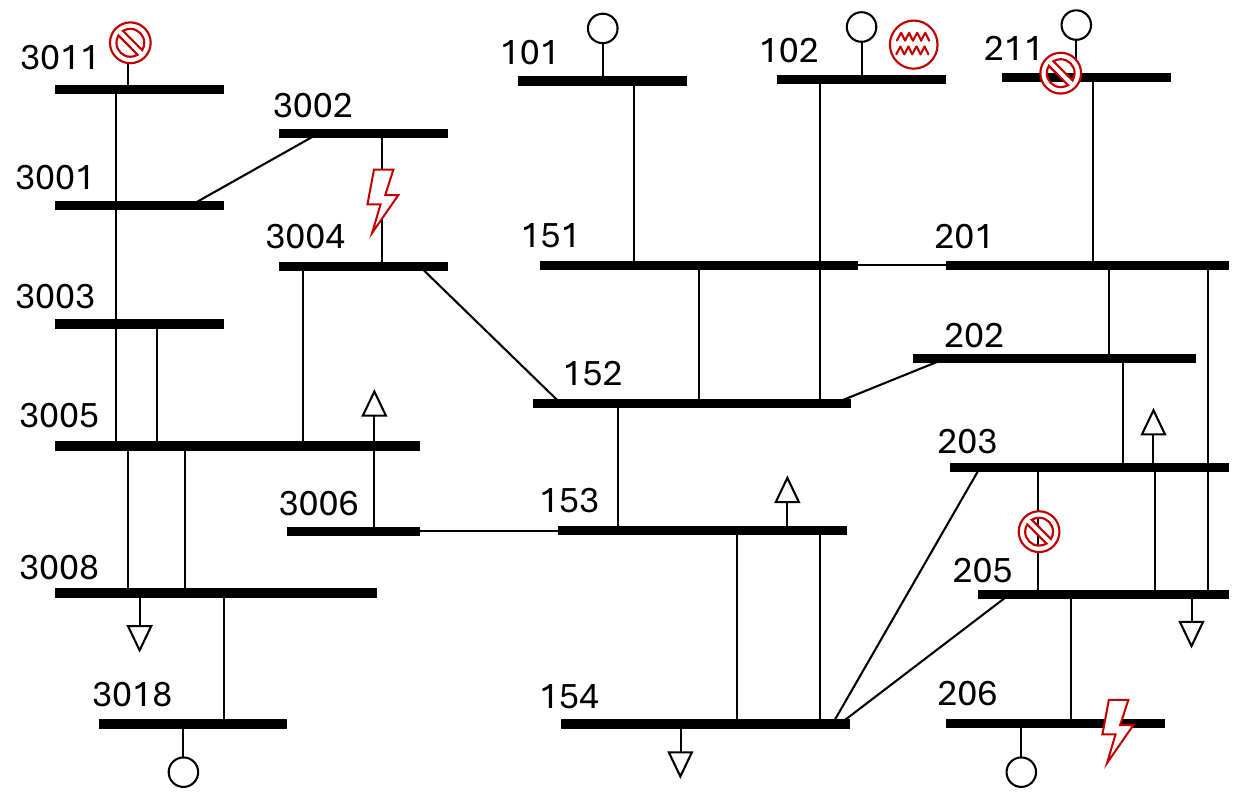}%
\small \caption{Visualization of the inter-connected electric grid network in PSML dataset\cite{zheng2022multi}, red symbols represent different fault types at different locations.}
\label{psml_network}
\end{figure}

Accurate fault diagnosis is crucial for enabling timely corrective action in electric grid operations. The Power Systems Machine Learning (PSML) dataset\cite{zheng2022multi} is a comprehensive dataset designed for testing fault type, fault location, and start time classification tasks. The data contains sensor readings from a network of phasor measurement units monitoring a 23-bus electric grid (see Fig. \ref{psml_network}). There are 91 feature variables (e.g., voltage, active power, reactive power, etc.) recorded for each time series, and these variables are likely to have complex cross-interactions due to the tightly coupled nature of the electric grid.

Each data sample contains disturbance events associated with any of the 5 fault types: bus fault, bus tripping, branch fault, branch tripping, or generator tripping. Moreover, the fault for each data sample can occur at any of the 53 possible locations  and 174 possible start-times, which makes this dataset particularly realistic and challenging for representation learning and classification. The data is also imbalanced, with roughly a 3:1 ratio between the majority and the minority fault class. The original dataset is already small-sized, thus, no further downsizing is required. We adopt the same train and test data used by the dataset's authors for benchmarking in \cite{zheng2022multi}.

\subsection{Benchmark Methods}
We evaluate our proposed method against a diverse set of baselines commonly used in time series classification benchmarks---including the  \textbf{classical} one-nearest neighbour (1-NN) DTW \cite{bagnall2018uea}, \textbf{standard} DL models (e.g., convolutional or recurent neural networks), and \textbf{state-of-the-art} DL methods (e.g. MLSTM-FCN \cite{karim2019multivariate}, TapNet \cite{zhang2020tapnet}, etc.) Among the state-of-the-art, we also assess self-supervised pre-training approaches (e.g. TS-TCC \cite{ijcai2021-324}, TF-C \cite{zhang2022self}) due to their established effectiveness in low labeled data settings.

Self-supervised learning enables the extraction of meaningful data representations by leveraging auxiliary learning tasks as supervisory signals, instead of manually annotated labels. Compared to fully supervised models, a smaller amount of labeled data is needed to fine-tune the learned representations for subsequent classification tasks. Existing self-supervised techniques can be broadly categorized into pre-text task learning, masked input reconstruction, and contrastive learning \cite{gui2024survey}. Pre-text task learning involves solving surrogate tasks (e.g., image jigsaw puzzles) to learn underlying feature representations. However, these tasks are mainly designed for computer vision and their generalizability is limited by the inductive biases of the pre-text task \cite{ijcai2021-324}. Masked input reconstruction predicts masked portions of data to capture feature dependencies, but it tends to predominantly learn granular, low-level patterns \cite{xie2023revealing}. In comparison, contrastive learning excels at capturing high-level, class-discriminative patterns, making it particularly suitable for classification tasks \cite{xie2023revealing}. Given its advantages, we focus on contrastive learning-based self-supervised benchmarks.

A brief commentary on the 7 baseline methods is given below, with additional emphasis on contrastive learning-based methods (i.e., TS-TCC \cite{ijcai2021-324}, TF-C \cite{zhang2022self}). 

\subsubsection{1-NN DTW} The 1-NN DTW \cite{bagnall2018uea} classifier operates by computing the similarity between a test sample and each training sample using a dynamic time warping-based distance metric. The test sample is then assigned the class label of the most similar training sample. In doing so, 1-NN DTW achieves instance discrimination by leveraging high-level sample-wise similarities. It is simple but effective, and continues to be a popularly benchmarked method as it can achieve best performance in some datasets \cite{zheng2022multi, karim2019multivariate, zhang2020tapnet}.
\subsubsection{Recurrent Neural Networks} Long short-term memory (LSTM) networks \cite{hochreiter1997long} and gated recurrent units (GRU) \cite{chung2014gru} are standard DL methods that are especially designed for time series learning. 
\subsubsection{MC-DCNN} Multi-channels deep convolution neural
network (MC-DCNN) \cite{zheng2014time} exploits the hierarchical feature extraction capabilities of deep CNNs to learn both high-level and low-level latent features in multivariate time series. Here, each channel handles a single dimension
of the multivariate time series and learns its latent features individually.

\subsubsection{MLSTM-FCN}
Multivariate long short-term memory-fully convolutional network (MLSTM-FCN) \cite{karim2019multivariate} combines LSTM and CNN to learn temporal relationships and cross-channel relationships between the different dimensions of a multivariate time series. Channel-wise dependencies in the CNN module are captured via a squeeze-and-excitation block, resulting in improved classification accuracies.  

\subsubsection{TapNet}
Time series attentional prototype network (TapNet)\cite{zhang2020tapnet} also combines LSTM and CNN for cross-dimensional learning. Additionally, it randomly permutes the time series dimensions prior to the CNN and incorporates an attention prototypical network to enhance the learning of cross-dimensional interactions amidst low labeled data.
\subsubsection{TS-TCC}
Time-Series representation
learning framework via Temporal and Contextual Contrasting (TS-TCC) \cite{ijcai2021-324} is specifically designed for time series contrastive learning. The original time series is augmented in two ways: weakly (jitter and scaling) and strongly (jitter and permutation). The original samples and their augmentations are fed into a three-block, 1D convolutional network-based encoder to extract latent temporal representations. In the subsequent temporal contrasting module, the temporal representations are refined through an autoregressive task that involves predicting the future timesteps of the weakly augmented representation using the strongly augmented one, and vice versa. Finally, contextual contrasting is applied to the temporal representations to learn further discriminative features for downstream classification. 

\subsubsection{TF-C}
Time series representation learning via Time-Frequency Consistency (TF-C) \cite{zhang2022self} is another state-of-the-art technique designed for time series contrastive learning. The basic premise is that time series representations can also be learned in the frequency domain, offering complementary insights into patterns that the time domain alone cannot fully capture. A three-layer 1D ResNet serves as the encoder for learning these representations. The representations are refined in the time-frequency latent space using contrastive learning, ensuring that the time- and frequency-based representations of a sample and its augmentation are close together, while dissimilar representations are pushed further apart. In addition to standard temporal augmentation techniques such as jittering and scaling, frequency-based augmentations are also introduced, where random frequency components are either removed or amplified in the frequency spectrum.

\subsection{Evaluation Metrics}
As recommended in \cite{zheng2022multi}, we use two performance metrics that are robust to potential class imbalances in the test data. For all classification tasks except for fault start-time detection, $balanced$ $accuracy$ \cite{sklearn-ba} is used for evaluation:
\begin{equation}
\begin{aligned}
\textit{\text{Balanced Accuracy}} &= \frac{\textit{\text{Sensitivity}} +  \textit{\text{Specificity}}}{2},\\[10pt]
\text{where} \quad 
\textit{\text{Sensitivity}} &= \frac{\textit{\text{True Positives}}}{\textit{\text{True Positives}} +  \textit{\text{False Negatives}}}, \\
\textit{\text{Specificity}} &= \frac{\textit{\text{True Negatives}}}{\textit{\text{True Negatives}} +  \textit{\text{False Positives}}}
\end{aligned}
\end{equation}

\noindent The value ranges from 0 to 1, and a higher value signifies better classification accuracy. It offers a more comprehensive view of the model performance by considering the correct classification of both majority and minority classes. As a further evaluation check, we also present the performance analysis with an additional metric, the $F1$  $score$ \cite{sklearn-f1}, in Appendix A.

For fault start-time detection, an ordinal classification task, we use the $macro-averaged\hspace{1mm}mean$ $absolute\hspace{1mm} error$ \textit{(MMAE)} \cite{baccianella2009evaluation, imlearn-mmae} instead for evaluation:

\begin{equation}
\begin{aligned}
\textit{\text{MMAE}} &= \frac{1}{P} \sum_{p=1}^{P} \text{MAE}_p, \\[10pt]
\text{where} \quad  \textit{\text{MAE}}_p &= \frac{1}{N_p} \sum_{n=1}^{N_p} \left| y^{(n)}_p - \hat{y}^{(n)}_p \right|,
\end{aligned}
\end{equation}

\noindent where \( y^{(n)}_p \) is the ground truth label of the \( n \)-th sample in class \( p \), \( \hat{y}^{(n)}_p \) is the corresponding prediction, \( \textit{\text{MAE}}_p \) is the mean absolute error for class \( p \), and \( \textit{\text{MMAE}} \) is the macro-average of all classes' \( \textit{\text{MAE}}_p \). Unlike $balanced$ $accuracy$,  \textit{MMAE} does not have a fixed upper value. Since neither too early nor too late fault detection are desirable, lower \textit{MMAE} values signify better classification performance as there is less divergence between the actual and predicted start-time \cite{zheng2022multi}.

\begin{table}{}
\newcolumntype{P}[1]{>{\centering\arraybackslash}p{#1}}
\renewcommand{\arraystretch}{1.3}
\scriptsize
\centering
\caption{ \centering Summary of selected hyperparameters.}
\vspace{0.1cm}
\label{hyperparameters}
\begin{tabular}{P{0.7cm}P{1cm}P{0.6cm}P{0.6cm}P{1.1cm}P{1cm}P{1cm}}

\hline                       \multirow{2}{0.7cm}{Module} & \multirow{2}{1cm}{\centering Hyper-parameters} & \multirow{2}{0.6cm}{QCAT Robot} & \multirow{2}{0.6cm}{PEMS Traffic} & \multicolumn{3}{c}{PSML Electric Grid} \\
\cline{5-7} 
 &  &  &  & Fault type & Location & Start-time \\
\hline                      
\multirow{2}{0.7cm}{\centering ITA}                             & $S$  & 12                     & 6                     & 6      & 6     & 6     \\
& \multirow{1}{1.1cm}{DTW type}  & shape & standard & standard      & shape & standard     \\
\hline                      
\multirow{3}{0.7cm}{\centering CTF}            & $R$ & 16 & 512 & 48 & 64 & 32 \\
 & $\alpha$ & 0.001 & 0.005 &  0.001 & 0.005     & 0.001\\
    & $\beta$ & 0.4 & 2.5 & 0.35     & 0.6    & 1.1     \\

\hline
\end{tabular}
\end{table}

\subsection{Experimental Settings}
Prior to creating the augmentations and training our model, the QCAT and PSML datasets are Z-score normalized \cite{sklearn-StandardScaler} to prevent exceptionally large data values from skewing distance-based metric learning. As the PEMS dataset contains occupancy rate values, which are already between 0 and 1, feature scaling is not required for it.

For each module of our proposed method, we report in Table \ref{hyperparameters} the optimal settings obtained by hyperparameter tuning. For the ITA module, $S\in\{6,12\}$ were considered as candidate mini-batch sizes, which balances good downstream classification performance (arising from pattern mixing and meaningful augmentations) against time needed for DTW-based searching of class prototypes. In terms of the DTW type used for the time series warping, both standard DTW \cite{sakoe1978dynamic} and shapeDTW \cite{zhao2018shapedtw} were considered.
In the CTF feature extractor module, the main hyperparameters are $R$, $\beta$, $\alpha$, and $\gamma$. Generally, the classification performance improves as the number of tensor components $R$ increases, until the optimal $R$ is reached (see Section \ref{sensitivity_analysis} for details). To search for the optimal $R$ in a manageable search space, we take the upper bound of $R$ to be the number of raw features $I$. We begin with initial guesses $R = I/2$ or $I/4$ and iterate with different values in increments of 16, making finer adjustments to smaller granularities if needed. Values for $\beta \in (0,5]$ were considered in increments of 0.5, and with finer granularity when necessary. For the learning rate $\eta$ and regularizer $\alpha$, standard values in $\{0.01,0.05, 0.001,0.005\}$ were evaluated. $\gamma$ is set to the batch size, making Eq. \ref{main_conloss} analogous to the standard NCE loss. Training was done for a maximum of 100 epochs with early stopping. The extracted features are passed to the downstream MLP classifier for the final classification. Experiments were repeated with 5 different seeds, and the mean results and standard deviation are reported.

For the benchmark models, a similar  search for optimal hyperparameters was conducted to obtain the best model performance. For the sake of brevity, we only discuss the common hyperparameters. For other custom hyperparameters, we follow the values recommended in the references cited for the respective benchmark models. Generally, candidate hidden layer sizes in $\{32, 64, 100, 128, 200, 256, 300 \}$, and where appropriate, number of hidden layers in $\{1, 2, 3 \}$  were also considered. For CNN-based architectures, additional hyperparameters such as the kernel size and number of filters in $\{ 3, 4, 5, 8 \}$ were considered. For all models, the range of candidate learning rates considered is same as that discussed for our proposed model. Furthermore, for a fair comparison, we evaluate TS-TCC and TF-C under the fine-tuning setting so that their performance can be enhanced with the use of labeled data.

\subsection{Results}
In this section, we provide a comprehensive analysis of the downstream classification results, including ablation studies on the contribution of contrastive learning and the proposed data augmentation strategy.

\begin{figure}[!t]
    \centering
    \subfloat[Samples 1 and 2 (top row, blue plots) have the same fault location label but different fault types, generator trip and bus trip, respectively. Thus, there is considerable intraclass variability within the same fault location class. The augmentation of Sample 1 (bottom left, red plot) captures temporal segments from Sample 2 (highlighted in yellow), illustrating the effectiveness of our augmentation strategy in achieving realistic intra-class pattern mixing.]{
        \begin{minipage}{0.9\linewidth}
            \centering
            \includegraphics[width=0.45\linewidth]{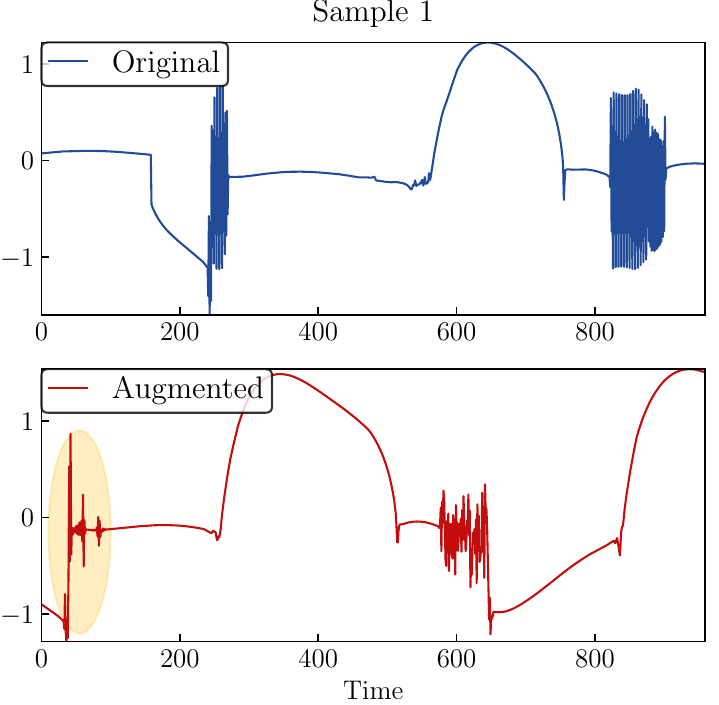}
            \hfill
            \includegraphics[width=0.45\linewidth]{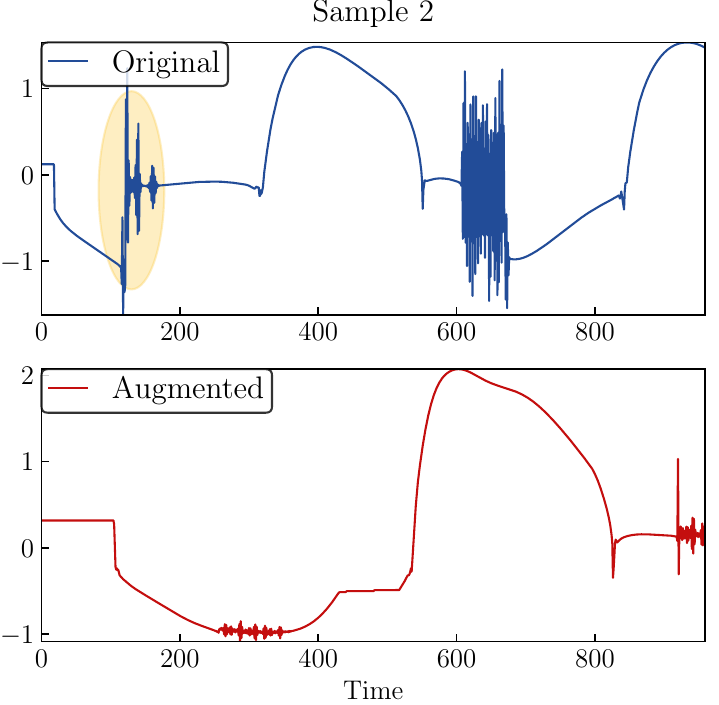}
        \end{minipage}
    }\\
    \vspace{0.5cm}
    \subfloat[Samples 2 and 3 (top row, blue plots) have the same fault location label but different fault types, bus trip and bus fault, respectively. Thus, there is considerable intraclass variability within the same fault location class. The augmentation of Sample 3 (bottom right, red plot) captures temporal segments from Sample 2 (highlighted in yellow), illustrating the effectiveness of our augmentation strategy in achieving realistic intra-class pattern mixing.]{
        \begin{minipage}{0.9\linewidth}
            \centering
            \includegraphics[width=0.45\linewidth]{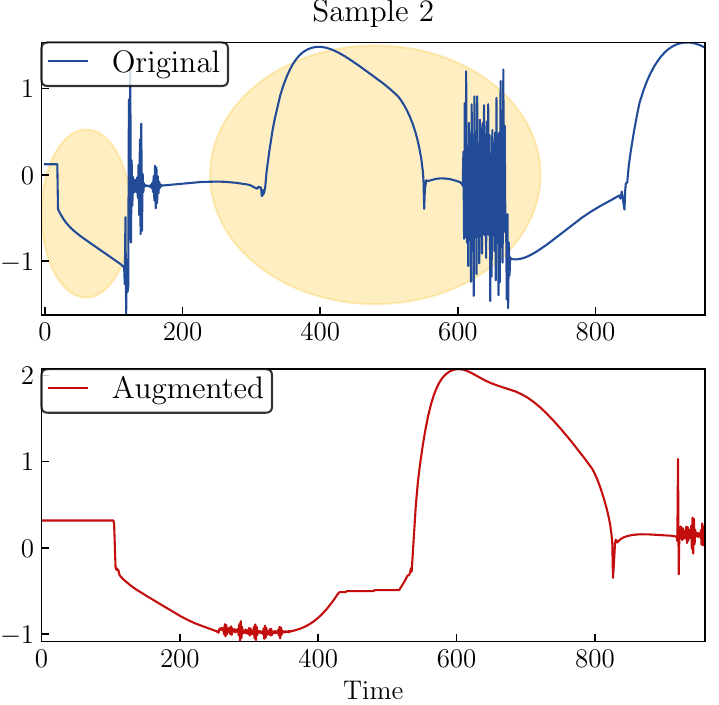}
            \hfill
            \includegraphics[width=0.45\linewidth]{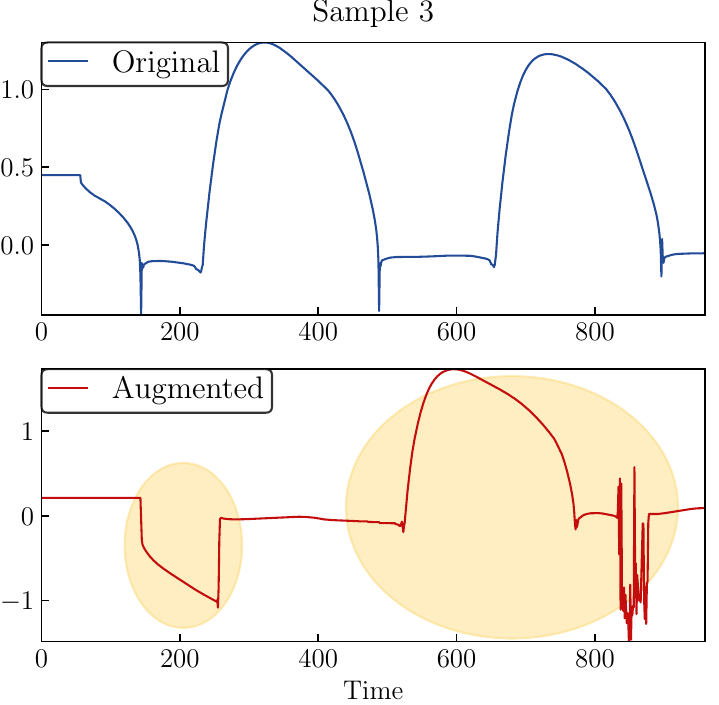}
        \end{minipage}
    }
    \caption{Data augmentation visualization, with the blue plots representing the original samples from the same location class but with different fault types. The red plots show the corresponding augmented samples, which exhibit realistic intra-class pattern mixing, as highlighted in yellow.}
    \label{fig_data_aug_viz}
\end{figure}

\begin{table*}[!h]
\newcolumntype{P}[1]{>{\centering\arraybackslash}p{#1}}
\renewcommand{\arraystretch}{1.3}
\scriptsize
\centering
\caption{\centering Performance comparison of proposed method against benchmarks.}
\vspace{0.1cm}
\label{comparison_results}
\begin{threeparttable}
\begin{tabular}{P{2.4cm}|P{1.75cm}P{1.65cm}P{1.65cm}P{1.65cm}P{1.8cm}}
\hline                         \multirow{3}{1cm}{\centering Method} & \multirow{2}{1.2cm}{\centering QCAT Robot} & \multirow{2}{1.2cm}{\centering PEMS Traffic} & \multicolumn{3}{c}{PSML Electric Grid} \\
\cline{4-6} 
  &  &  & Fault type\tnote{ \textdaggerdbl} & Location\tnote{ \textdaggerdbl} & Start-time\tnote{ \textdaggerdbl} \\
\cline{2-5}\cline{6-6}                                           
 & \multicolumn{4}{c|}{Balanced Accuracy ($\uparrow$)}    & \textit{MMAE} ($\downarrow$) \\
\cline{1-5} \cline{6-6}                                           
 1-NN DTW \cite{bagnall2018uea}\tnote{\textdagger} & \textbf{0.913} & 0.709 & 0.598 & 0.474 & 53.709 \\
 LSTM \cite{hochreiter1997long} & 0.644 ± 0.228
 & 0.863 ±0.025
 & 0.544 ± 0.049
 & 0.248 ± 0.043
  & 56.434 ± 2.851
   \\
 GRU \cite{chung2014gru} & 0.721 ± 0.256
 & 0.814 ± 0.036
& 0.653 ± 0.029 & 0.332 ± 0.062 & 55.550 ± 2.090
     \\
                                MC-DCNN \cite{zheng2014time} & 0.867± 0.013
 & 0.789 ± 0.037 &  0.726 ± 0.019 & 0.437 ± 0.030 & 38.107 ± 5.675
 \\
                                MLSTM-FCN \cite{karim2019multivariate} & 0.865 ± 0.032
 &  0.781 ± 0.071
 &  0.742 ± 0.029
  &  0.285 ± 0.023
 & 31.873 ± 5.400
    \\
TapNet \cite{zhang2020tapnet}  & 0.856 ± 0.006 & 0.700 ± 0.097 & 0.653 ± 0.018 & 0.397 ± 0.065   & 58.251 ± 1.974 \\
TS-TCC \cite{ijcai2021-324} & 0.866 ± 0.038 & 0.785 ± 0.016 & 0.674 ± 0.023 & 0.376 ± 0.030 & 30.368 ± 3.051 \\
TF-C \cite{zhang2022self} & 0.866 ± 0.020 & 0.635 ± 0.023 & 0.606 ± 0.024 & 0.218 ± 0.027 & 34.650 ± 1.987 \\
ITA-CTF (proposed) & 0.895 ± 0.016
 & \textbf{0.875 ± 0.005} & \textbf{0.745 ± 0.022} & \textbf{0.602 ± 0.018} & \textbf{26.632 ± 1.817} \\
\hline  
\end{tabular}
\begin{tablenotes}
\item[\textdagger] The 1-NN DTW results do not change across trials, thus the standard deviation is zero.
\item[\textdaggerdbl] Results from \cite{zheng2022multi} were reported for 1-NN DTW to TapNet as they were on par or better than our trials.
\end{tablenotes}
\end{threeparttable}
\vspace{-1.2 em}
\end{table*}

Before presenting the results, we briefly examine the generated data augmentations to illustrate the pattern mixing and time warping effects introduced by the proposed data augmentation technique. We use the most challenging dataset, PSML Electric Grid, as an example to highlight the intra-class variability  in the original data and visualize the proposed data augmentations. Fig. \ref{fig_data_aug_viz} shows a pairwise comparison of 3 random samples taken from the same fault location class (Bus 3011), but each having a different fault type---namely, generator trip, bus trip, and bus fault. Consequently, the sensor signals of these samples exhibit considerable intra-class variability as seen from the blue-colored plots. Consequently, the sensor signals of these samples (blue plots) exhibit considerable intra-class variability for the same fault location class, making the location classification task highly challenging. Our proposed data augmentations are designed to emphasize intra-class variations.  As shown in the augmentations of the samples (red plots), feature patterns from other samples (highlighted in yellow) have been blended into the augmented sample, resulting in impactful augmentations with greater ``teaching'' ability. By feeding in such targeted and efficient augmentations for contrastive learning, we can induce the CTF feature extractor to explicitly learn the intra-class variability despite having limited training data.

\subsubsection{Classification Results}
Table \ref{comparison_results} compares our proposed method and established benchmarks on 3 challenging datasets from different domains, covering a total of 5 classification tasks. Our proposed method outperforms the benchmarks on 4 out of 5 classification tasks. For the QCAT Robot terrain type classification, our method is outperformed by 1-NN DTW, but remains a close second-best. 

Interestingly, while some benchmarks showed strong, second-best performance, none performed as consistently well as the proposed method across all classification tasks---likely due to the significantly different feature learning requirements. For example, 1-NN DTW and MC-DCNN performed well for terrain type and fault location classification, owing to their respective abilities for instance discrimination and learning spatial relationships. In particular, 1-NN DTW's outperformance on terrain classification may be attributed to its strength in high-level sample-wise similarity matching, which aligns well with the QCAT dataset’s relatively lower complexity (e.g., fewer interconnected sensor channels and target classes). However, both 1-NN DTW and MC-DCNN fared worse for fault start-time classification, where learning in-depth temporal features is critical. In this vein, TS-TCC, which is adept at learning temporal features among limited data, performs well on the start-time detection, but fares worse with fault location classification when complex feature learning beyond temporal features is needed. Surprisingly, TF-C struggles across most tasks and datasets, performing the worst on location classification. This could be because frequency-based contrastive learning may not always be meaningful for time series across all domain applications, and random frequency perturbation-based augmentations could potentially disrupt important dependencies. Moreover, the 1D convolution-based encoders of both TS-TCC and TF-C may be inadequate for capturing cross-dimensional interactions inherent in complex multi-dimensional time series data. Finally, MLSTM-FCN is designed to learn both sensor-wise and temporal relationships, but its performance suffers when faced with a large number of raw feature dimensions and small training data, as evidenced by the PEMS Traffic and fault location classification performance.

In comparison, our method is consistently competitive, ranking among the top-performers for all classification tasks. Its performance also remains robust on highly complex tasks with a large number of classes, such as fault location and start-time classification. This is a strong testament to the importance of (\textit{i}) simultaneously encoding sensor factors, temporal factors, and their cross-interactions, and (\textit{ii}) having targeted and informative  augmentations during CL to guide the fine-grained learning of complex features in small training datasets.

\begin{figure}[!t]
\vspace{-1.2 em}
\centering
\subfloat[MLSTM-FCN]{\includegraphics[width=0.45\linewidth]{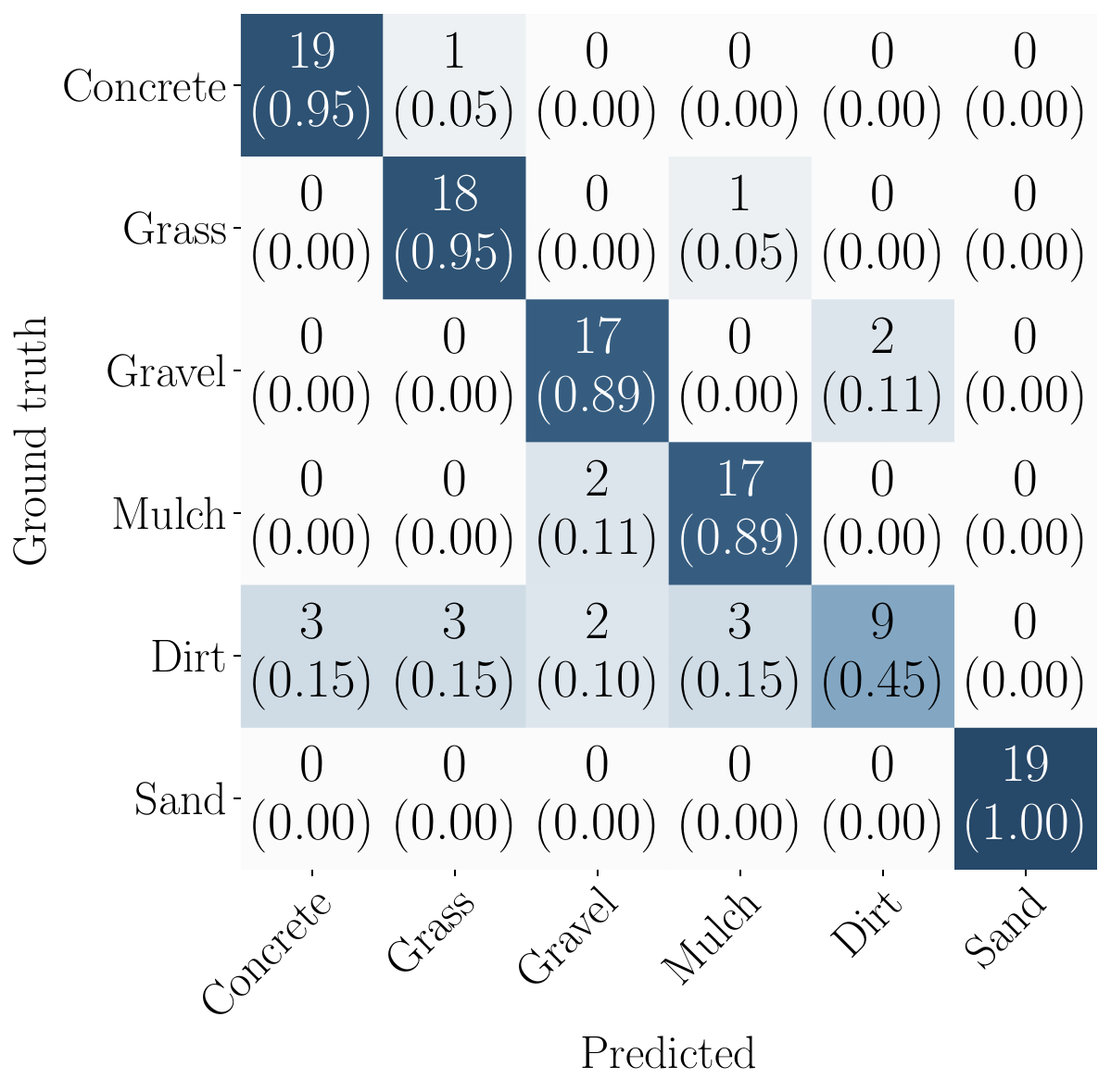}%
\label{MLSTM-FCN_performance}}
\hfil 
\subfloat[ITA-CTF (Proposed)]{\includegraphics[width=0.45\linewidth]{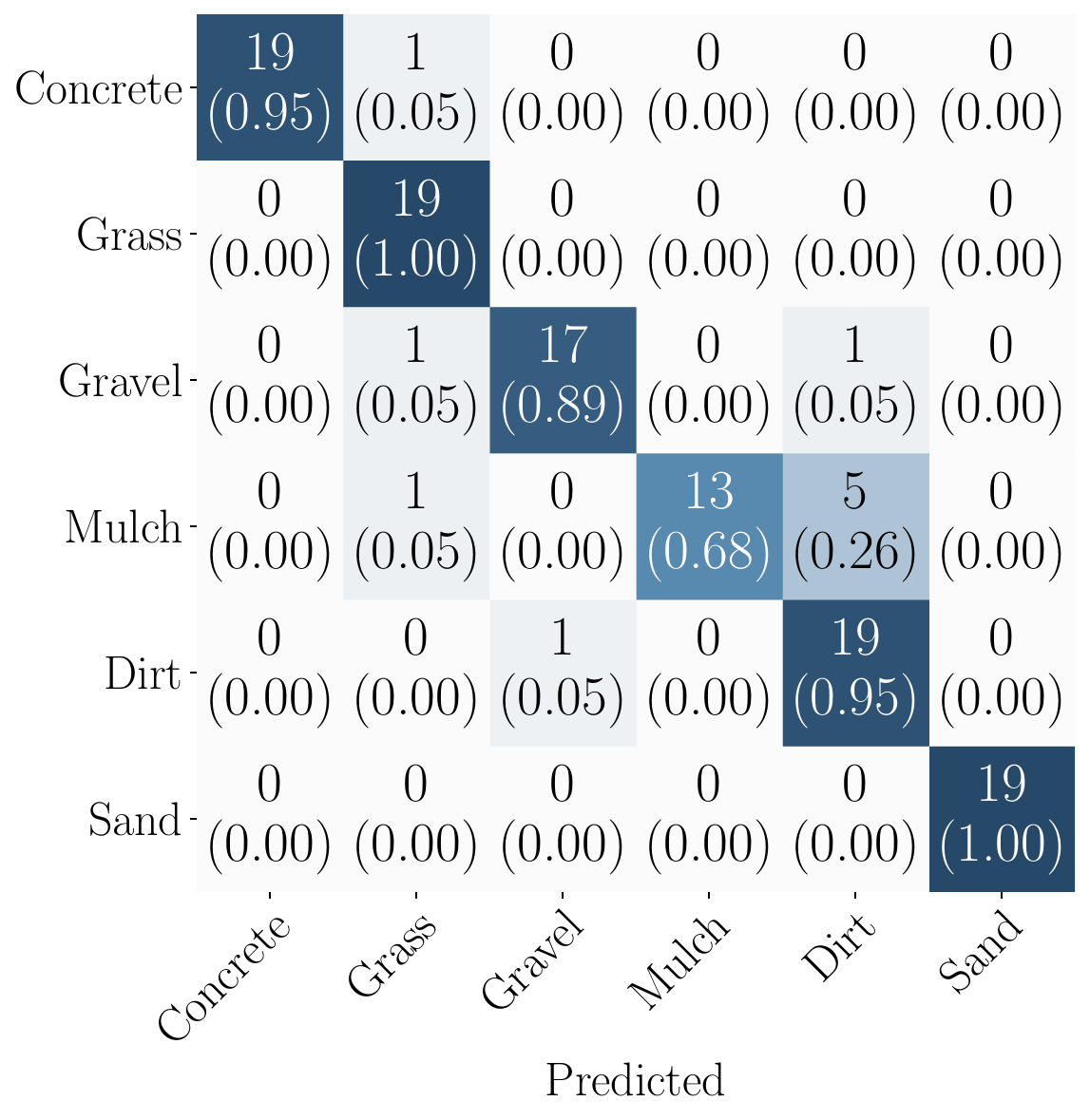}%
\label{ITA-CTF_performance}}
\small \caption{Confusion matrix highlighting fine-grained classification ability of proposed model over benchmark model.}
\label{confusion_matrix_fig}
\end{figure}

For a deeper analysis of fine-grained learning capabilities under low-data settings, we compare the in-depth classification performance of our proposed model and MLSTM-FCN (a DL benchmark specially designed for multi-dimensional time series). Taking the QCAT Robot terrain type classification task as an example,  Fig. \ref{confusion_matrix_fig} plots the class-wise prediction performance of the two models. As evidenced in Fig. \ref{confusion_matrix_fig}\subref{MLSTM-FCN_performance}, MLSTM-FCN struggles to classify the highly challenging `dirt' class as it is very similar to other classes, such as `mulch' or `grass', in terms of surface mechanical properties. Out of the 20 `dirt' class samples, MLSTM-FCN misclassifies 11 of them, achieving a recall of 45\%.  For 25\% of the samples, it also misclassifies the `dirt' class as classes with markedly different mechanical properties such as `concrete' and `gravel'. This misclassification can pose serious challenges for smooth navigation of robotic systems. In comparison, the proposed ITA-CTF model, plotted in Fig. \ref{confusion_matrix_fig}\subref{ITA-CTF_performance}, classifies the `dirt' class with a impressive recall of 95\%. One limitation is that it misclassifies  26\% of the `mulch' class samples as `dirt'. However, this has far less practical consequences due to the close mechanical similarities between the two classes.

\subsubsection{Ablation Studies}

We evaluate the impact of the different components of our proposed method in enhancing the quality of the learnt representations for classification. First, we assess the role of CL by comparing the performance of a standard TF model without any contrastive loss (i.e., only reconstruction loss and regularization) against our proposed method. Next, we investigate the role of meaningful data augmentations in enhancing CL's effectiveness by comparing the proposed augmentation strategy against four popular augmentation techniques for time series: jittering \cite{um2017data}, permutation \cite{um2017data}, time-warping \cite{um2017data}, and mixup \cite{zhang2018mixup}. For mixup, we take guidance from Zhang \textit{et al.} \cite{zhang2018mixup} for the recommended augmentation parameters. For the remaining techniques, we follow Um \textit{et al.} \cite{um2017data}'s recommendations. Jittering creates weak augmentations that preserve the overall patterns of the original signal by just adding random Gaussian noise with a mean of 0 and standard deviation of 0.03. In contrast, permutation creates stronger augmentations by dividing the original time series into a number of smaller segments and randomly shuffling these segments. We let the number of segments to cut the time series vary randomly from 1 to 5. Similarly, time-warping generates augmentations by applying non-linear transformations to the time series signal of a sample, along the time dimension. Lastly, mixup  creates augmentations by linearly combining two random samples with a weight $\lambda \in [0, 1]$, where $\lambda$ is sampled from the $\operatorname{Beta}(\theta, \theta)$ distribution, with $\theta \in (0, \infty)$. We set $\theta = 2$ for a more even mixing of the samples.

The results of the ablation studies are reported in Table \ref{ablation_studies}. First, it is evident that the standard TF model (without CL) can already achieve competitive performance, even surpassing several DL models from Table \ref{comparison_results}. This is likely due  to the explicit accounting of cross-interactions of sensor and temporal factors in the TF model, which is crucial for learning representations from  networked, time series. Next, we note that effectiveness of the CL addition to the TF model is dependent on the quality of the data augmentation. The CTF model with standard data augmentations could not achieve consistent performance improvements across all datasets. It is possible that the mild augmentations produced by jittering are not meaningfully capturing the intra-class variability present in the small but complex datasets. Meanwhile, the stronger augmentations produced by permutations, time-warping, and mixup could be naively disrupting temporal dependencies, resulting in unrealistic patterns for the class distribution. This may cause a  performance deterioration, especially if the classification task relies heavily on temporal dependencies (e.g., start-time detection). 

In contrast, the proposed ``soft'' class prototype-based ITA module is task-agnostic, and yields meaningfully pattern-mixed and time-warped augmentations. This enables us to effectively harness the powerful, class-aware representation learning ability of CL, while also inducing learning of the intra-class variations within classes.  Compared to standard TF, we observe strong performance improvements, ranging from 1.2\% to 18.7\%, across all datasets. 
\begin{table}[!t]
\newcolumntype{P}[1]{>{\centering\arraybackslash}p{#1}}
\renewcommand{\arraystretch}{1.3}
\scriptsize
\centering
\caption{\centering Ablation studies.}
\vspace{0.1cm}
\label{ablation_studies}
\begin{threeparttable}
\begin{tabular}{P{2.23cm}|P{1cm}P{0.9cm}P{1.1cm}P{1cm}P{1.1cm}}
\hline                         \multirow{3}{1cm}{\centering Components} & \multirow{2}{1cm}{\centering QCAT Robot} & \multirow{2}{1cm}{\centering PEMS Traffic} & \multicolumn{3}{c}{PSML Electric Grid} \\
\cline{4-6} 
  &  &  & Fault type & Location & Start-time \\
\cline{2-5}\cline{6-6}                                           
 & \multicolumn{4}{c|}{Balanced Accuracy ($\uparrow$)}    & \textit{MMAE} ($\downarrow$) \\
\cline{1-5} \cline{6-6}                                           
 Standard TF & 0.884\tiny±0.013 & 0.845\tiny±0.015 & 0.725\tiny±0.035 & 0.507\tiny±0.027 & 29.402\tiny±2.350 \\
Jittering + CTF  & 0.886\tiny±0.014
 & 0.861\tiny±0.018
 & 0.726\tiny±0.031
 & 0.514\tiny±0.047
  & 29.647\tiny±1.624
   \\
Permutation + CTF & 0.874\tiny±0.020
 & 0.862\tiny±0.017
& 0.733±\tiny0.023 & 0.476\tiny±0.021 & 30.344\tiny±1.657
     \\
Time-warping + CTF & 0.886\tiny±0.024 & 0.859\tiny±0.019 & 0.734±\tiny±0.022 & 0.527\tiny±0.015 & 29.816\tiny±1.593
     \\    
Mixup + CTF & 0.882\tiny±0.010 & 0.856\tiny±0.009 & 0.725±\tiny±0.040 & 0.484\tiny±0.025 & 29.025\tiny±2.133
     \\      
     
                                ITA + CTF (proposed) & \textbf{0.895\tiny±0.016}
 & \textbf{0.875\tiny±0.005} & \textbf{0.745\tiny±0.022} & \textbf{0.602\tiny±0.018} & \textbf{26.632\tiny±1.817}
 \\
\hline  
\end{tabular}
\end{threeparttable}
\vspace{-1.2 em}
\end{table}

\begin{figure*}[!h]
\centering
\subfloat[QCAT Robot]{\includegraphics[width=0.6\linewidth]{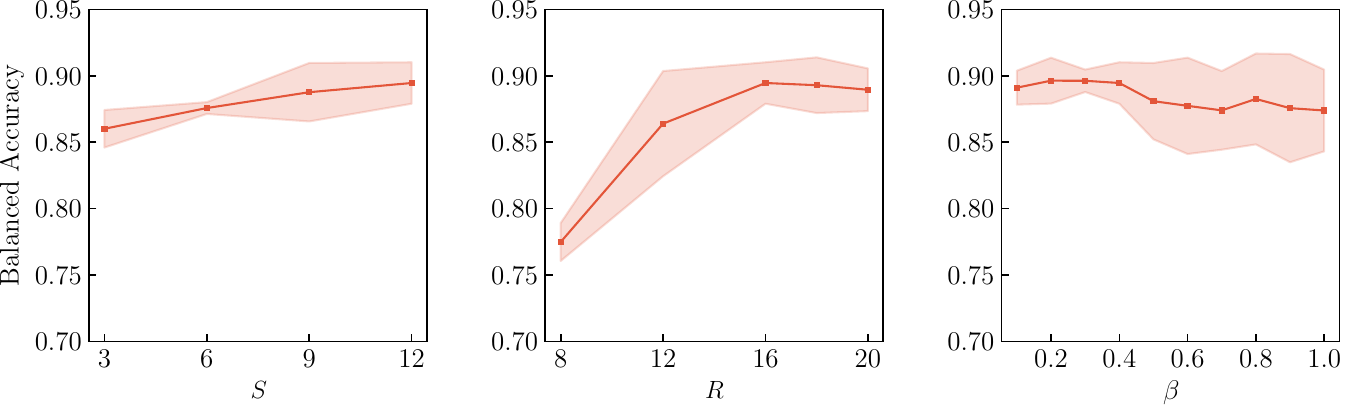}%
\label{QCAT_sensitivity}}
\vspace{7mm}
\subfloat[PEMS Traffic]
{\includegraphics[width=0.6\linewidth]{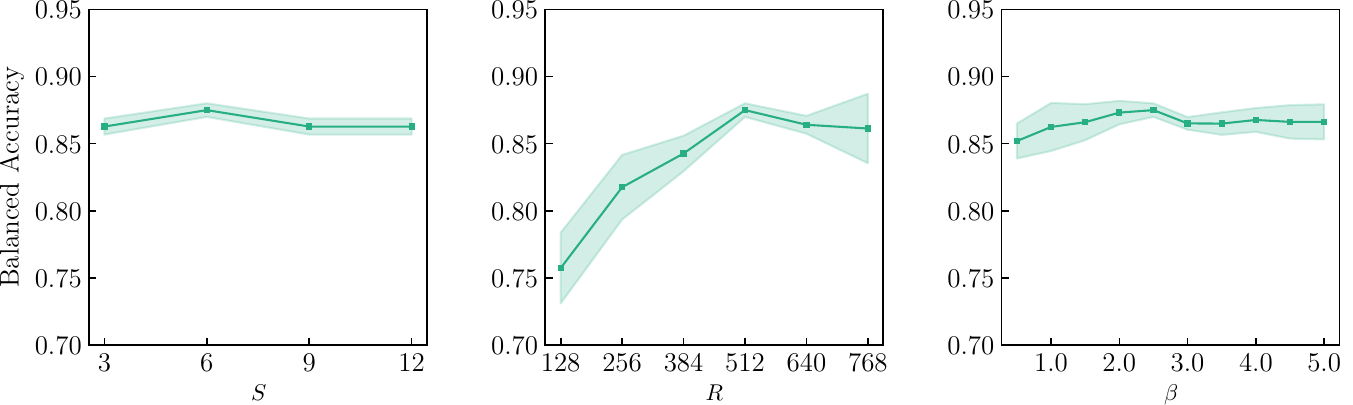}%
\label{PEMS_sensitivity}}
\vspace{7mm}
\subfloat[PSML Location]{\includegraphics[width=0.6\linewidth]{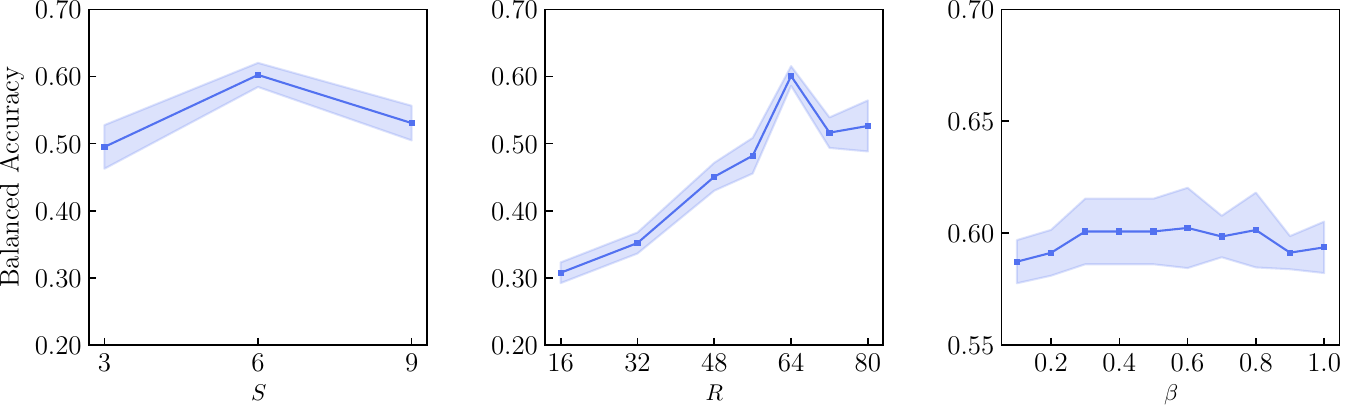}%
\label{PSML_Location_sensitivity}}
\caption{Performance analysis on the choice of $S$, $R$, and $\beta$ on classification accuracy.}
\vspace{-1.3em}
\label{fig_sensitivity}
\end{figure*}

\subsubsection{Hyperparameter Sensitivity and Performance Analysis}
\label{sensitivity_analysis}
We conduct a performance analysis to evaluate the impact of three critical hyperparameters on the classification accuracy: the mini-batch size $S$ for dynamically determining a ``soft'' class prototype, the number of components $R$ and the contrastive loss weight $\beta$. The classification performance trends for a range of candidate $S$, $R$, and $\beta$ values are plotted in Fig. \ref{fig_sensitivity} for the three datasets. For the PSML dataset, we focus on the location classification task as a challenging representative example.

The choice of $S$ impacts downstream classification performance through its effect on the  quality of data augmentations generated. To recap, the ITA module uses a sampling-based approach, selecting from a mini-batch of size $S \ll N$, to dynamically determine the ``soft'' class prototypes required for generating data augmentations. Generally, if $S$ is too small (e.g., $S=3$, as shown in Fig. \ref{fig_sensitivity}), the mini-batch size may be insufficient to identify a class prototype that effectively discriminates between classes. However, as $S$ increases, the generated class prototypes may converge to become more deterministic, thereby reducing their ability to produce a greater variety of intra-class variations.   Moreover, computation time for solving Eq. \ref{class_prototype_eqn} with a larger batch of samples can be prohibitive, especially for datasets such as PSML Electric Grid, which have both high feature dimensions and long sequence lengths. Thus, the practical choice of $S$ balances the classification performance and the computation time. In our experiments, with the exception of the QCAT Robot dataset, the optimal size $S$ for the best  classification performance is 6. For the QCAT Robot dataset, classification performance plateaus at a larger size, around 9 to 12.

The choice of $R$ in TF-based models is crucial for classification performance as it directly determines the model size (i.e., the number of parameters) and the representation learning capacity. As seen in Fig. \ref{fig_sensitivity}, increasing the number of decomposition components $R$ generally improves classification performance, as a higher $R$ enables the TF process to capture more latent explanatory components of the data that aid classification. However, there is typically an optimal $R$ value, beyond which classification performance tends to decline, possibly due to overfitting. Model overfitting occurs when models fit to noise and idiosyncrasies of the training data rather than learning meaningful representations, leading to  poor generalization and classification performance on test data. As summarized in Table \ref{hyperparameters}, the optimal $R$ for each classification task is dataset-specific but is consistently much smaller than the number of feature dimensions, suggesting that an effective model captures only the most salient latent structures while leaving out irrelevant noise in the dataset.

Nonetheless, overfitting risks are especially relevant in low training data regimes. We address this risk, first, from an architectural design perspective by proposing a TF-based approach over supervised DL models. Their lower model complexity, combined with an inherent ability to capture important structural dependencies, make TF models better suited for mitigating overfitting risks when learning from multi-dimensional data with limited training samples. Second, we apply regularization \cite{calvetti2003tikhonov} during model training to constrain the model parameter values, preventing them from growing excessively large and overfitting to irrelevant noise in the training data. By limiting the parameter size, regularization encourages the model to focus on the most important patterns in the data, rather than memorizing specific examples. In particular, we incorporate an L2 regularization term $\mathcal{L}_{reg}$ into the overall loss function (Eq. \ref{overall_loss}) to penalize large parameter values and constrain the growth of parameters $\mathbf{Z}$, $\tilde{\mathbf{Z}}$, $\mathbf{A}$, and $\mathbf{B}$. 
The hyperparameter $\alpha$ in Eq. \ref{overall_loss} controls the strength of the regularization, and we find optimal $\alpha$ values to be 0.001 or 0.005, depending on the dataset. With these well-designed strategies to mitigate overfitting, our model generally outperforms baselines in test classification, even with limited training data.

Finally, the hyperparameter $\beta$ controls the relative contribution of the reconstruction loss $\mathcal{L}_{rec}$ and the contrastive loss $\mathcal{L}_{con}$ in the overall loss function (Eq. \ref{overall_loss}). As previously reported in the ablation study (Table \ref{ablation_studies}), classification performance benefits from incorporating the contrastive loss $\mathcal{L}_{con}$ (i.e., $\beta > 0$) compared to using only the reconstruction loss in standard TF. Fig. \ref{fig_sensitivity} shows that there is generally an optimal $\beta$ that yields peak classification performance by effectively balancing the reconstruction and contrastive losses. However, it is worth noting that our model exhibits considerable robustness to the choice of $\beta$, as classification performance remains relatively stable across different $\beta$ values. Suboptimal $\beta$ values do not cause a drastic decline in performance.

\begin{table*}{}
\newcolumntype{P}[1]{>{\centering\arraybackslash}p{#1}}
\renewcommand{\arraystretch}{1.3}
\scriptsize
\centering
\caption{\centering Scalability analysis for ITA module with varying number of channels and sequence length.}
\vspace{0.1cm}
\label{scalability_analysis}


\begin{tabular}{P{2.2cm}P{0.8cm}P{2.9cm}P{3cm}}
\hline 
Dataset name & Channels ($I$) & Sequence length ($J$) & Average time per augmented sample (seconds)  \\
\hline 
PEMS Traffic  & 963   & 144  & 0.274  \\

QCAT Robot    & 22   & 662 \textcolor{gray}{ (4.6 $\times $ PEMS)}  & 1.693 \textcolor{gray}{ (6.2 $\times $ PEMS)} \\

PSML Location  & 91   & 960 \textcolor{gray}{ (6.7 $\times $ PEMS)}  & 3.544 \textcolor{gray}{ (12.9 $\times $ PEMS)} \\
\hline
\end{tabular}

\parbox{\linewidth}{\vspace{1mm} \scriptsize \hspace{1cm} Note: For a fair comparison, we fix the mini-batch size to $S =6$ for all datasets and use standard DTW for augmentation.}
\end{table*}

\begin{table*}[h]
\newcolumntype{P}[1]{>{\centering\arraybackslash}p{#1}}
\renewcommand{\arraystretch}{1.3}
\centering
\scriptsize
\caption{Runtime efficiency comparison of proposed method against benchmarks.}
\vspace{0.1cm}
\label{runtime_efficiency_analysis}
\begin{tabular}{P{2cm}P{1.6cm}P{2cm}P{2.8cm}P{2.1cm}P{2cm}}
\hline
Model & Dataset & Average train time per epoch (seconds) & Average train time per sample (seconds) & Average test time per sample (seconds) & Classification accuracy \\
\hline
\multirow{3}{*}{ITA-CTF (proposed)} 
    & QCAT & 1.226 (CTF) & 58.578 (ITA) + 0.160 (CTF) & $1.896 \times 10^{-6}$ & \textbf{0.895} \\
    & PEMS & 2.802 (CTF) & 0.274 (ITA) + 0.840 (CTF) & $5.653 \times 10^{-6}$ & \textbf{0.875} \\
    & PSML Location & 1.505 (CTF) & 29.936 (ITA) +  0.274 (CTF)  & $1.158 \times 10^{-5}$  & \textbf{0.602} \\
\hline
\multirow{3}{*}{MLSTM-FCN} 
    & QCAT & 0.768 & 0.209 & 0.022 & 0.865 \\
    & PEMS & 0.422 & 0.218 & 0.015 & 0.781 \\
    & PSML Location & 0.860 & 0.122 & 0.024 & 0.285 \\
\hline
\multirow{3}{*}{MC-DCNN} 
    & QCAT & 12.178 & 1.324 & 0.775 & 0.867 \\
    & PEMS & 1.965 & 1.015 & 0.084 & 0.789 \\
    & PSML Location & 18.820 & 2.145 & 1.429 & 0.437 \\
\hline
\multirow{3}{*}{TS-TCC} 
    & QCAT & 0.847 & 0.161 & $4.236 \times 10^{-5}$ & 0.866 \\
    & PEMS & 1.206 & 0.623 & $1.131 \times 10^{-4}$ & 0.785 \\
    & PSML Location & 1.477 & 0.337 & $6.429 \times 10^{-5}$ & 0.376 \\
\hline
\end{tabular}
\end{table*}

\subsubsection{Scalability and Complexity Analysis}
\label{complexity_analysis}

We provide a comprehensive analysis of the computational efficiency and  scalability of the proposed ITA-CTF framework. The framework essentially consists of two core modules. The ITA module for generating intelligent augmentations has a complexity of $\mathcal{O}\hspace{0.03cm}(SJ^2I)$ per batch, primarily governed by the $\mathcal{O}\hspace{0.03cm}(J^2I)$ complexity of dynamic time warping. The CTF module for feature extraction has a complexity of $\mathcal{O}\hspace{0.03cm}(NJIR)$ per epoch, determined by the alternating least squares algorithm.

For the scalability analysis, we focus on the ITA module, which has the higher computational complexity of $\mathcal{O}\hspace{0.03cm}(J^2I)$. For a fixed mini-batch size $S$, we investigate the computational time required to generate an augmentation sample as the number of channels $I$ and sequence length $J$ of the sample increase. As shown in Table \ref{scalability_analysis}, the primary determinant of computation time is sequence length $J$. This is evident from the PEMS Traffic dataset, which, despite having the largest number of channels $I$, has the lowest computation time due to its shorter sequence length. However, when comparing the QCAT Robot and PSML Location datasets to PEMS, we observe that the increase in computation time with sequence length is much less than the theoretically expected quadratic growth (see Table \ref{scalability_analysis} for details). These performance gains are likely due to practical optimizations or speedups that occur even with standard computational resources.

We also compare the runtime efficiency of our full ITA-CTF framework with competitive DL baselines. Specifically, Table \ref{runtime_efficiency_analysis} reports the training and test time of our model alongside the best-performing baselines from our classification experiments: MLSTM-FCN, MC-DCNN, and TS-TCC. For a fair comparison, we include the one-time computation time of generating intelligent augmentations under the train time per sample. Excluding this offline augmentation process, we find that the per-epoch training time of our feature extraction module, CTF, is highly competitive with existing baselines while achieving significantly higher classification accuracy.  More importantly, once training is complete, the test time on the trained model is negligible---on the order of $10^{-5}$ seconds---making it highly practical for real-time classification tasks. We attribute the low test time to the simple architectural structure of the CTF feature extractor and MLP classifier, where the burden of learning robust features was primarily addressed through the intelligent augmentations and the enhanced contrastive tensor factorization optimization objective.

\subsubsection{Limitations and Future Directions}
\label{limitations_future_dir}

While our method achieves notable performance across a variety of industrial applications in practical low-data settings, there are two potential areas for improvement that can further boost its practical value in real-world industrial applications. First, there is scope for further optimizing the training time. In our current model, the online inference time is fast---on the order of $10^{-5}$ seconds---making it well-suited for real-time inference. However, the offline computation time for generating ``soft'' class prototype-based data augmentations can be further optimized, especially for time series with longer sequence lengths. Thus, in future work or for practical deployment, we can explore advanced DTW algorithms, such as parallelizable DTW \cite{tralie2020exact} for GPU acceleration, or lower bound-based techniques \cite{tan2019elastic} for early pruning of unpromising matches to reduce unnecessary computations.

Second, the fault localization task proved particularly challenging in our experiments due to the complex combination of strong spatio-temporal dependencies, a large number of target classes, and limited training data. While our model outperforms all benchmarks, there is certainly scope to further improve the localization performance. To this end, leveraging advanced learning paradigms such as meta-learning\cite{finn2017model} offers a promising direction to boost our model’s classification performance on few-shot tasks while maintaining data efficiency. For instance, recent studies have proposed contrastive meta-learning to derive robust representations from partial observations \cite{jelley2023contrastive}, and have exploited latent feature spaces to generate informative augmentations for contrastive learning \cite{li2022metaug}. However, most existing meta-learning strategies have been developed for computer vision tasks, and adapting them to complex, multi-dimensional time series presents open challenges. Moreover, designing meta-learning techniques that leverage knowledge learned from different tasks (e.g., fault type \textit{and} location classification) to quickly fine-tune new tasks with limited data (e.g., new fault locations) offers a promising avenue for extending our model.

\section{Conclusion}
\label{conclusion}
In this work, we proposed an innovative, data-efficient framework, Intelligently Augmented Contrastive Tensor Factorization (ITA-CTF), for fine-grained learning of complex features in multi-dimensional time series. Our method fills an important practical gap for the classification of real-world time series, where the need for learning complex features, such as multi-factor dependencies and intra-class variations, are often hampered by low labeled training data availability. In our extensive experiments spanning five different classification tasks, our model consistently ranked among the top-performers. It also significantly outperformed benchmark models on highly complex tasks with a large number of classes but limited training samples, such as fault location and start-time classification. This is a strong testament to  the CTF module's ability to sieve out intra-class variations present in real-world time series data, and seek out important invariant properties (e.g., sensor factors, temporal factors, and their joint interactions) that characterize multi-dimensional time series. This contrastive learning process was significantly strengthened by the ITA module's  intelligently pattern-mixed augmentations that efficiently  highlight realistic intra-class patterns to the CTF module despite the small original training data. Although our proposed method focuses on industrial problems, its domain- and task-agnostic nature makes it easily generalizable to other domains, such as smart healthcare, where learning from multi-dimensional sensor data with limited labeled data is often critical.

\bibliographystyle{IEEEtran}
\bibliography{mybib.bib}

\appendices
\renewcommand{\thetable}{A.\arabic{table}}  
\section{Classification performance with F1 score}
\label{appendix_A}

%
\setcounter{table}{0} 

As an additional evaluation, the classification performance is assessed using the weighted F1 score \cite{sklearn-f1}, which also accounts for class imbalance. The results are reported in Table \ref{comparison_results_f1score}.

\begin{table}[!h]
\newcolumntype{P}[1]{>{\centering\arraybackslash}p{#1}}
\renewcommand{\arraystretch}{1.3}
\scriptsize
\centering
\caption{\centering Performance comparison of proposed method against benchmarks using F1 Score ($\uparrow$).}
\vspace{0.1cm}
\label{comparison_results_f1score}
\begin{threeparttable}
\begin{tabular}{P{2.2cm}|P{1cm}P{1cm}P{1.2cm}P{1cm}}
\hline                         \multirow{2}{1cm}{\centering Method} & \multirow{2}{1.2cm}{\centering QCAT Robot} & \multirow{2}{1.2cm}{\centering PEMS Traffic} & \multicolumn{2}{c}{PSML Electric Grid} \\
\cline{4-5} 
  &  &  & Fault type & Location \\
\cline{1-5}                                           
 1-NN DTW \cite{bagnall2018uea}\tnote{\textdagger} & \textbf{0.912} & 0.704 & 0.636 & 0.578 \\
 LSTM \cite{hochreiter1997long} & 0.641\tiny ±0.221
 & 0.858\tiny ±0.045
 & 0.606\tiny ±0.068
 & 0.432\tiny ±0.027
  \\
 GRU \cite{chung2014gru} & 0.683\tiny ±0.284
 & 0.843\tiny ±0.029
 & 0.714\tiny ±0.066
 & 0.534\tiny ±0.033
     \\
                                MC-DCNN \cite{zheng2014time} & 0.863\tiny ±0.013
 & 0.789\tiny ±0.039
 & 0.756\tiny ±0.024
 & 0.550\tiny ±0.036
 \\
                                MLSTM-FCN \cite{karim2019multivariate} & 0.866\tiny ±0.028
 & 0.749\tiny ±0.094
 & 0.719\tiny ±0.054
 & 0.406\tiny ±0.044\\
TapNet \cite{zhang2020tapnet}  & 0.856\tiny ±0.009
 & 0.698\tiny ±0.106
 & 0.651\tiny ±0.048
 & 0.468\tiny ±0.021  \\
TS-TCC \cite{ijcai2021-324} & 0.863\tiny ±0.041
 & 0.783\tiny ±0.018
 & 0.698\tiny ±0.016
 & 0.519\tiny ±0.022
  \\
TF-C \cite{zhang2022self}  & 0.863\tiny ±0.021
 & 0.642\tiny ±0.024
 & 0.625\tiny ±0.043
 & 0.345\tiny ±0.025  \\
\textit{ITA-CTF (proposed)} & 0.894\tiny ±0.016
 & \textbf{0.875\tiny ±0.005}
 & \textbf{0.780\tiny ±0.021}
 & \textbf{0.649\tiny ±0.014}  \\
\hline  
\end{tabular}
\begin{tablenotes}
\item[\textdagger] The 1-NN DTW results do not change across trials, thus the standard deviation is zero.
\end{tablenotes}
\end{threeparttable}
\vspace{-1.2 em}
\end{table}

\
\section{Interpretability Analysis}
Tensor factorization provides \textit{mechanistic interpretability} by explicitly multi-dimensional data into factor matrices that capture the contributions of different factors (e.g., sensor, temporal) and a coefficient matrix representing cross-factor interactions. For the interpretability analysis, we focus on the terrain classification task of the QCAT dataset due to its smaller data dimensions and fewer target classes, which make the visualizations easier to interpret.  However, the general principles of interpreting the learned factor and coefficient matrices with respect to the latent components apply to all datasets and tasks. In Fig. \ref{factor_matrices}(a), we visualize the sensor factor matrix $\mathbf{A} \in \mathbb{R}^{I \times R}$ to assess the relative contribution of each of the $I$ sensors to the $R$ latent components. Similarly, in Fig. \ref{factor_matrices}(b), the temporal factor matrix $\mathbf{B} \in \mathbb{R}^{J \times R}$ illustrates how each latent component evolves across the $J$ time steps. For instance, by examining the first column (corresponding to component $R=1$), we observe that sensors 12 to 16 and time steps up to 126 are most prominently activated, highlighting their strong association with this component.

\renewcommand{\thefigure}{B.\arabic{figure}} 
\begin{figure}[!h]
    \centering
    \setcounter{figure}{0}
    \subfloat[Sensor Factor Matrix \textbf{A}]{%
        \includegraphics[width=0.3\textwidth]{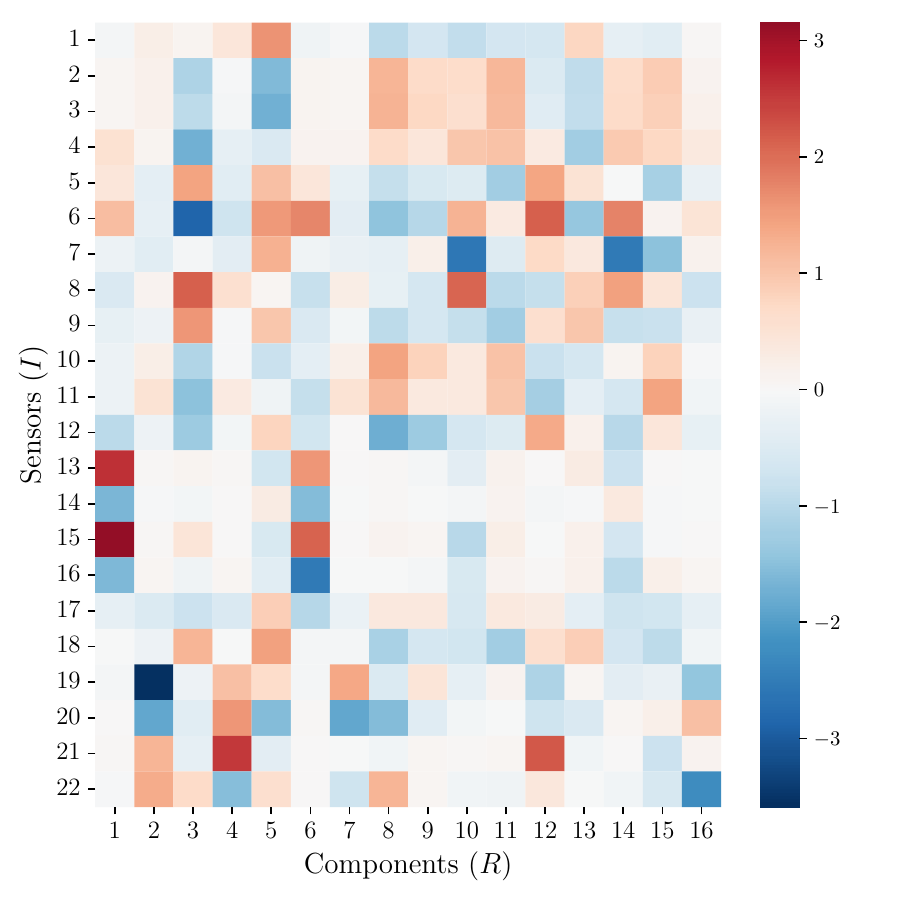}%
    }
    \hspace{0.5cm}
    \subfloat[Temporal Factor Matrix \textbf{B}]{%
        \includegraphics[width=0.3\textwidth]{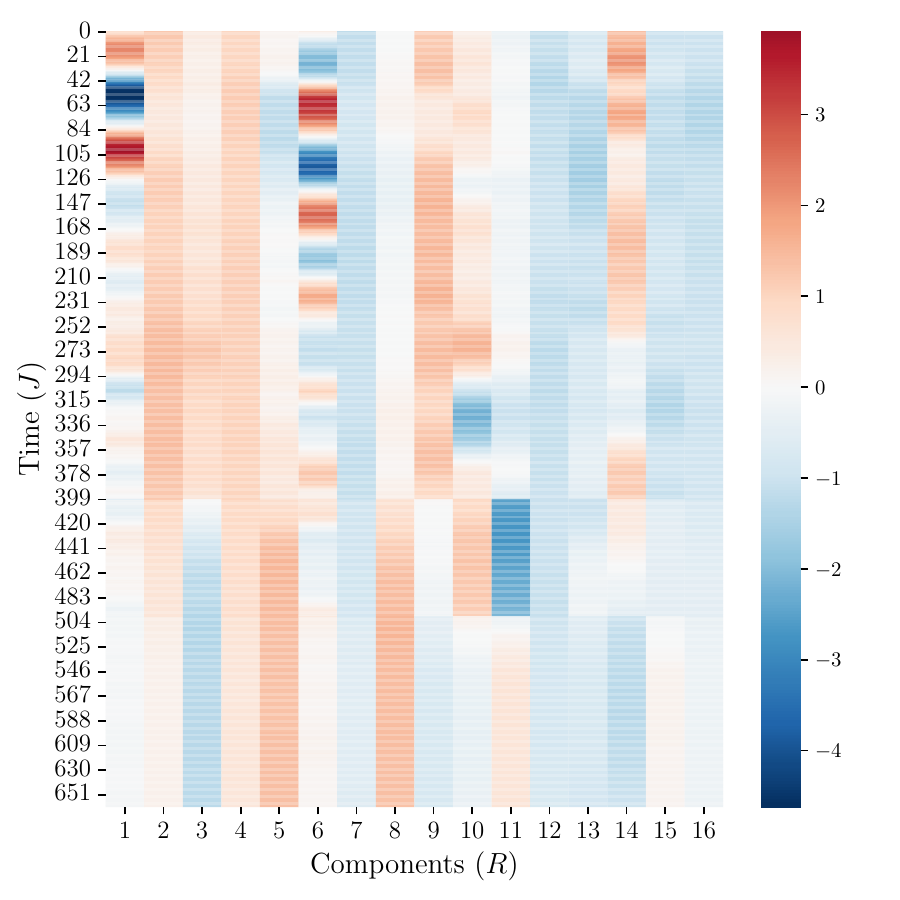}%
    }
    \caption{Visualization of learnt factor matrices.}
    \label{factor_matrices}
\end{figure}

\begin{figure*}[!h]
    \centering
    \subfloat[Concrete]{%
        \includegraphics[width=0.3\textwidth]{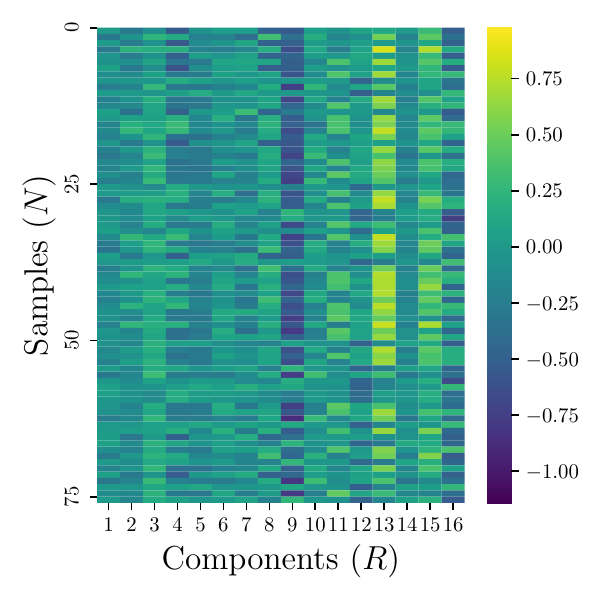}%
    }
    \hspace{0.1cm}
    \subfloat[Grass]{%
        \includegraphics[width=0.3\textwidth]{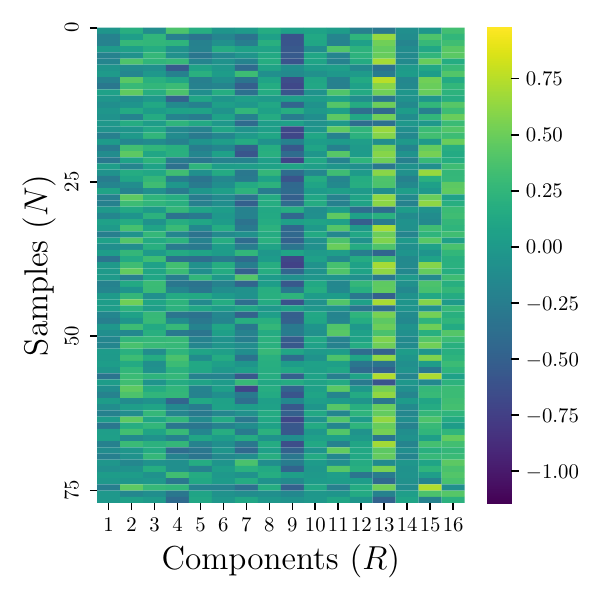}%
    }
    \hspace{0.1cm}
    \subfloat[Gravel]{%
        \includegraphics[width=0.3\textwidth]{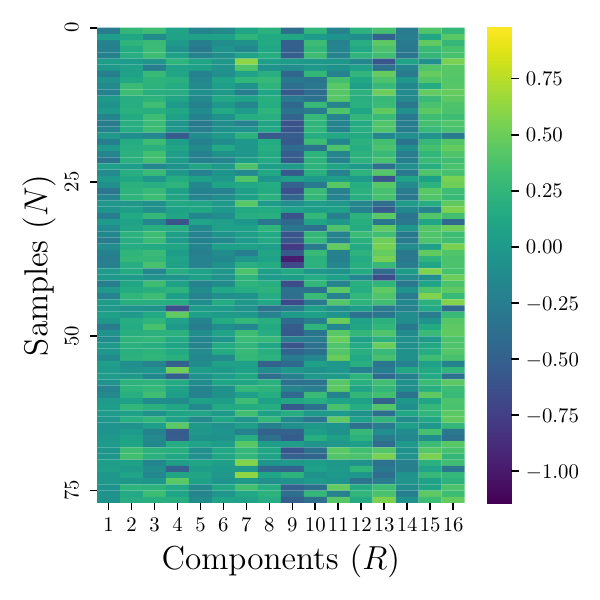}%
    }
    \vfill 
    \subfloat[Mulch]{%
        \includegraphics[width=0.3\textwidth]{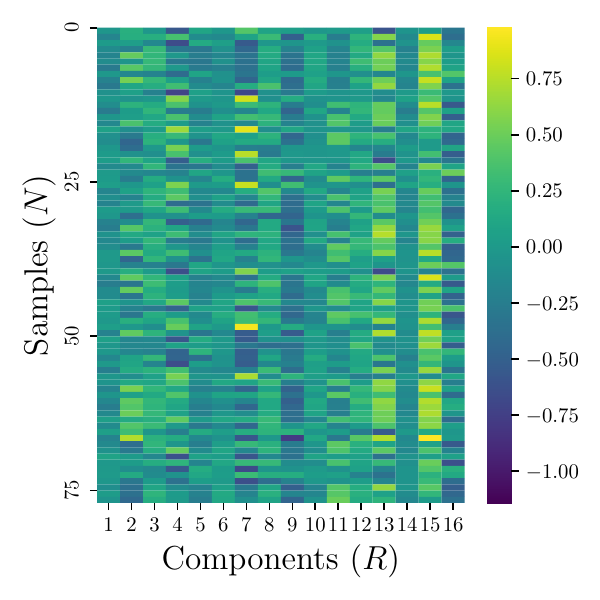}%
    }
    \hspace{0.1cm}
    \subfloat[Dirt]{%
        \includegraphics[width=0.3\textwidth]{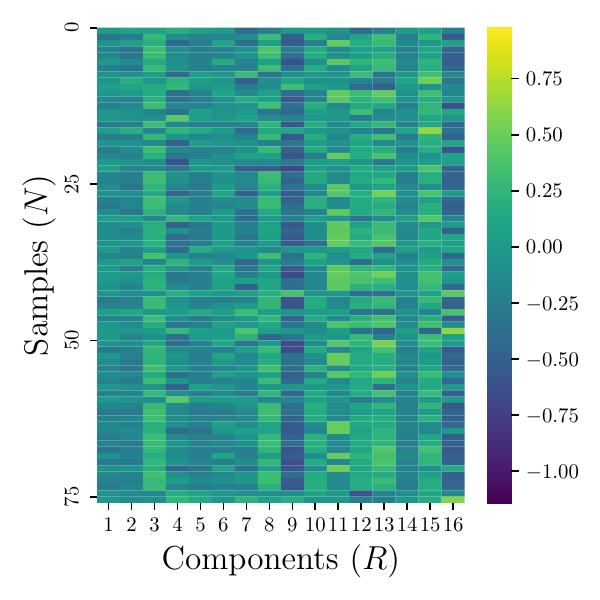}%
    }
    \hspace{0.1cm}
    \subfloat[Sand]{%
        \includegraphics[width=0.3\textwidth]{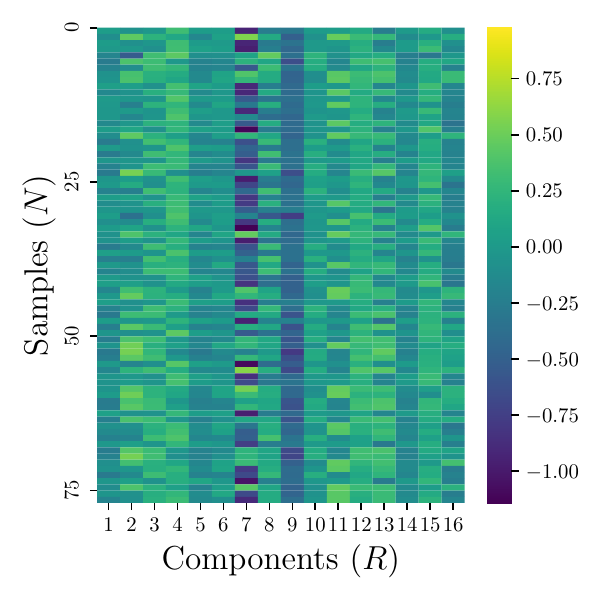}%
    }
    \caption{Visualization of learnt coefficient matrix (feature extractions) for samples from each class, encapsulating sensor factors, temporal factors, and cross-factor dependencies.}
    \label{coefficient_matrices}
\end{figure*}

Next, in Fig. \ref{coefficient_matrices}, we visualize the coefficient matrix (i.e., the learned feature representations that capture cross-factor interactions) for each terrain class. Two key observations emerges. First, within each terrain class, the feature representations of the samples are generally similar, reflecting shared characteristics within the class, which aids in classification. Second, across classes, each class exhibits distinct signatures with different activated components. For example, in the `concrete' class, component $R = 13$ is activated, whereas in the `sand' class, component $R = 7$ is activated instead, potentially highlighting the contrasting mechanical properties of the two classes.

\end{document}